\documentclass[conference]{IEEEtran}
\usepackage{times}

\usepackage[numbers]{natbib}
\usepackage{multicol}
\usepackage[bookmarks=true]{hyperref}

\usepackage{amssymb} % for checkmarks
\usepackage{gensymb} % for degree symbol
\usepackage{xcolor}
\usepackage{mathtools}
\usepackage{graphicx}  % figures
\usepackage{subfig}  % to make subfigures
\usepackage{multirow}  % tables: spanning text across multiple columns
\usepackage{makecell}  % tables: inserting linebreaks within a cell
\newcolumntype{?}{!{\vrule width 1pt}}  % tables: boldface horizontal dividers
\usepackage[ruled,vlined]{algorithm2e} % for algorithms

\definecolor{deeppink}{rgb}{1.0, 0.08, 0.58}
\usepackage{hyperref}
\hypersetup{
  colorlinks = true, %colours links instead of ugly boxes
  urlcolor   = deeppink, %colour for external hyperlinks
  linkcolor  = blue, %colour of internal links
  citecolor  = red %colour of citations
}
\usepackage{pifont} % for checkmarks and crosses
\newcommand{\cmark}{\ding{51}}%
\newcommand{\xmark}{\ding{55}}%
\usepackage{color, soul} % This allows us to highlight text that you want your co-authors to review
\newcommand{\Go}{\mathcal{G}}  % original constraint graph
\newcommand{\Ge}{\overline{\mathcal{G}}}  % extended constraint graph
\newcommand{\X}{\mathbf{X}}

\newcommand{\R}{R}
\newcommand{\V}{\mathbf{\Omega}}
\newcommand{\No}{\mathbf{N}}  % original node
\newcommand{\Ne}{\overline{\mathbf{N}}}  % extended node

\newcommand{\set}{\mathcal{S}}

\newcommand{\tlas}{\theta}
\newcommand{\I}{\mathbf{I}}
\newcommand{\dd}{\mathbf{D}}

\newcommand{\pdet}{P_\mathrm{det}}

\newcommand{\website}{https://siddancha.github.io/projects/active-safety-envelopes-with-guarantees}

\renewcommand\hl[1]{#1} % Uncomment this to turn off highlighting

\begin{document}

\title{Active Safety Envelopes using Light Curtains with Probabilistic Guarantees}

% You will get a Paper-ID when submitting a pdf file to the conference system
% \author{Author Names Omitted for Anonymous Review. Paper-ID 157}

% \author{
%   \authorblockN{Siddharth Ancha}
%   \authorblockA{Carnegie Mellon University\\
%   Pittsburgh, PA 15213\\
%   \texttt{\small sancha@cs.cmu.edu}}
  
%   \and

%   \authorblockN{Gaurav Pathak}
%   \authorblockA{Carnegie Mellon University\\
%   Pittsburgh, PA 15213\\
%   \texttt{\small gauravp@andrew.cmu.edu }}

%   \and

%   \authorblockN{Srinivasa G. Narasimhan}
%   \authorblockA{Carnegie Mellon University\\
%   Pittsburgh, PA 15213\\
%   \texttt{\small srinivas@andrew.cmu.edu }}

%   \and

%   \authorblockN{David Held}
%   \authorblockA{Carnegie Mellon University\\
%   Pittsburgh, PA 15213\\
%   \texttt{\small dheld@andrew.cmu.edu}}
% }

\author{
  \newcommand{\width}{2.5em}
  \authorblockN{
    Siddharth Ancha
    \hspace{\width}
    Gaurav Pathak
    \hspace{\width}
    Srinivasa G. Narasimhan
    \hspace{\width}
    David Held
  }
  \authorblockA{
    Carnegie Mellon University, Pittsburgh PA 15213, USA\\
    \{\texttt{sancha, gauravp, srinivas, dheld}\}\texttt{@andrew.cmu.edu}}\\

    Website: \href{\website}{\small \texttt{\website}}
}

% avoiding spaces at the end of the author lines is not a problem with
% conference papers because we don't use \thanks or \IEEEmembership

%Homer Simpson\authorrefmark{2},
%Montgomery Scott\authorrefmark{3} and
%Eldon Tyrell\authorrefmark{4}}
%Telephone: (800) 555--1212, Fax: (888) 555--1212}
%\authorblockA{\authorrefmark{4}Tyrell Inc., 123 Replicant Street, Los Angeles, California 90210--4321}}

\maketitle

% \begin{strip}
%   \centering
%   \small
%   Website\footnote{asdfsa}: \href{\website}{\texttt{\website}}
% \end{strip}

\begin{abstract}
To safely navigate unknown environments, robots must accurately perceive dynamic obstacles. Instead of directly measuring the scene depth with a LiDAR sensor, we explore the use of a much cheaper and higher resolution sensor: \textit{programmable light curtains}. Light curtains are controllable depth sensors that sense only along a surface that a user selects.  We use light curtains to estimate the \textit{safety envelope} of a scene: a hypothetical surface that separates the robot from all obstacles. We show that generating light curtains that sense \textit{random} locations (from a particular distribution) can quickly discover the safety envelope for scenes with unknown objects. Importantly, we produce theoretical safety guarantees on the probability of detecting an obstacle using random curtains. We combine random curtains with a machine learning based model that forecasts and tracks the motion of the safety envelope efficiently. Our method accurately estimates safety envelopes while providing probabilistic safety guarantees that can be used to certify the efficacy of a robot perception system to detect and avoid dynamic obstacles. We evaluate our approach in a simulated urban driving environment and a real-world environment with moving pedestrians using a light curtain device and show that we can estimate safety envelopes efficiently and effectively.\footnote{\hl{Please see our \mbox{\href{\website}{\small project website}} for (1) a web-based demo of random curtain analysis, (2) videos showing qualitative results of our method and (3) source code.}}
\end{abstract}

\IEEEpeerreviewmaketitle

%%%%%%%%%%%%%%%%%%%%%%%%%%%%%%%%%%%%%%%%%%%%%%%%%%%%%%%%%%%%%%%%%%%%%%%%%%%%%%%%
% Sections
%%%%%%%%%%%%%%%%%%%%%%%%%%%%%%%%%%%%%%%%%%%%%%%%%%%%%%%%%%%%%%%%%%%%%%%%%%%%%%%%

\section{Introduction}
\label{sec:introduction}

Consider a robot navigating in an unknown environment. The environment may contain objects that are arbitrarily distributed, whose motion is haphazard, and that may enter and leave the environment in an undetermined manner. This situation is commonly encountered in a variety of robotics tasks such as autonomous driving, indoor and outdoor robot navigation, mobile robotics, and robot delivery. How do we ensure that the robot moves safely in this environment and avoids collision with obstacles whose locations are unknown a priori? What guarantees can we provide about its perception system being able to discover these obstacles?

Given a LiDAR \hl{sensor}, the locations of obstacles can be computed from the captured point cloud; however, LiDARs are typically expensive and low-resolution. Cameras are cheaper and high-resolution and 2D depth maps of the environment can be predicted from the images. However, depth estimation from camera images is prone to errors and does not guarantee safety.

An alternative approach is to use \textit{active perception}~\cite{bajcsy1988active,bajcsy2018revisiting}, where only the important and required parts of the scene are accurately sensed, by actively guiding a controllable sensor in an intelligent manner. 
%We follow the latter approach and propose to directly estimate the safety envelope. By doing so, we avoid both the complexity of predicting the full depth map, as well as the post-processing necessary to compute the envelope.
Specifically, a programmable light curtain~\cite{wang2018programmable,bartels2019agile,ancha20eccv} is a light-weight \textit{controllable} sensor that detects objects intersecting any user-specified 2D vertically ruled surface (or a `curtain'). Because they use an ordinary rolling shutter camera, \hl{light curtains combine the best of both worlds of passive cameras (high spatial-temporal resolution and lower cost) and LiDARs (accurate detection along the 2D curtain and robustness to scattered media like smoke/fog)}.
%First, light curtains can measure depth at very high spatial resolutions. Second, they can be cheaply constructed (a current lab-built prototype costs \$1000, and the price is expected to reduce significantly in production). 
%Light curtains can be adaptively placed in the environment to sense a scene intelligently and efficiently. 
%However, they raise two novel and challenging questions that need to be addressed: Where do we place the curtains or safety envelopes without {\it a priori} knowledge of objects in the scene? How do we evolve those envelopes over time to capture a dynamic scene?

%present a challenge to the user who must decide where to place the light curtain; only the locations where the curtain gets placed are sensed.

%The objective of this work is to accurately estimate safety envelopes of a scene, across a given sequence of sensor data. 

%Because light curtains can only be placed along a user-specified vertically ruled surface, 
In this work, we propose to use light curtains to estimate the ``safety envelope" of a scene.  We define the safety envelope as an imaginary, vertically ruled surface that separates the robot from all obstacles in the scene. The region between the envelope and the robot is free space and is safe for the robot to occupy without colliding with any objects. Furthermore, the safety envelope ``hugs" the \hl{closest} object surfaces to maximize the amount of free space between the robot and the envelope. \hl{More formally, we define a safety envelope as a 1D depth map that is computed from a full 2D depth map by selecting the closest depth value along each column of the 2D depth map (ignoring points on the ground or above a maximal height).}
%More formally, given a 2D depth map of the scene (projected from a birds-eye view), the safety envelope corresponds to the closest point along each column to the robot (between a minimum ground height and a maximum ceiling height). 
%The safety envelope must move over time with the motion of the obstacles in the scene.
As long as the robot never intersects the safety envelope, it is guaranteed to not collide with any obstacle.

%However, uniformly sensing the full depth of a scene, which is the dominant paradigm for depth perception, may be unnecessary and inefficient. 

% \begin{figure*}[h]
%     \centering
%     \includegraphics[draft,width=0.6\textwidth,trim=0 0 0 0]{figures/network_architecture.png}
%     \caption{\sid{Pull figure: a cool example of our method working. Show hugging due to main curtain, re-detecting object due to random curtain.}}
%     \label{fig:pull}
% \end{figure*}

%Hence, they can be used to perform active perception tasks. In this work, we use light curtains to actively estimate the safety envelope of the scene. \sid{Dave: may be you can phrase the following point better}. Note that we do not treat safety envelope estimation using light curtains as a replacement to full scene perception using traditional sensors. Instead, we consider them to be an important supplement for ensuring safety and obstacle avoidance by robot perception systems.

Realizing this concept requires addressing two novel and challenging questions: First, where do we place the curtains without {\it a priori} knowledge of objects in the scene? The light curtain will only sense the parts of the scene where the curtain is placed.  Second, how do we evolve these curtains over time to capture dynamic objects?
%The goal of this paper is to develop a method to use a light curtain to detect the safety envelope of a scene with objects whose locations are unknown. 
One approach is to place light curtains at random locations in the unknown scene. Previous work~\cite{bartels2019agile} has empirically shown that random light curtains can quickly discover unknown objects. In this work, we develop a systematic framework to generate random curtains that respect the physical constraints of the light curtain device. Importantly, we develop a method that produces theoretical guarantees on the probability of random curtains (from a given distribution) to detect unknown objects in the environment and discover the safety envelope. Such safety guarantees could be used to certify the efficacy of a robot perception system to detect and avoid obstacles.

Once a part of the safety envelope (such as an object's surface) is discovered, it may be inefficient to keep exploring the scene randomly. Instead, a better strategy is to forecast how the identified safety envelope will move in the next timestep and track it by sensing at the predicted location. We achieve this by training a neural network to forecast the \hl{position of the envelope in the next timestep using previous light curtain measurements}.  
However, it is difficult to provide theoretical guarantees for such learning-based systems. 
We overcome this challenge by combining the deep neural network with random light curtain placements. Using this combination,  we are able to estimate the safety envelope efficiently, while furnishing probabilistic guarantees for discovering unknown obstacles. 
%Both the forecasting model and random curtains are crucial components for accurately identifying the safety envelope of a scene, towards building a robot perception system that can safely and efficiently detect obstacles to avoid.
Our contributions are:

% Machine learning based methods could be trained to forecast and track the envelope. However, it is often hard to develop theoretical guarantees on the performance of machine learning models, especially neural networks. On the other hand, our method is able to provide probabilistic guarantees for safety envelope discovery by random curtains. 

%In this work, we leverage the best of both worlds.

\begin{enumerate}
    \item We develop a systematic framework to generate random curtains that respect the physical constraints of the light curtain device, by extending the ``light curtain constraint graph'' introduced in prior work~\cite{ancha20eccv} (Sec.~\ref{sec:extended-graph},~\ref{sec:random-curtain-analysis:sampling-random-curtain}).
    \item We develop a dynamic-programming based approach to produce theoretical safety guarantees on the probability of random curtains discovering unknown objects in the environment (Sec.~\ref{sec:random-curtain-analysis:theoretical-analysis},~\ref{sec:experiments:analysis}).
    \item We combine random light curtains with a machine learning based forecasting model to efficiently estimate safety envelopes (Sec.~\ref{sec:learning-forecasting}).
    \item We evaluate our approach on (1) a simulated autonomous driving environment, and (2) a real-world environment with moving pedestrians. We empirically demonstrate that our approach consistently outperforms multiple baselines and ablation conditions (Sec.~\ref{sec:experiments:depth-tracking}).
\end{enumerate}

\section{Related Work}

\subsection{Active perception and light curtains}

Active perception involves actively controlling a sensor for improved perception~\cite{bajcsy1988active,bajcsy2018revisiting}, such as controlling camera parameters~\cite{bajcsy1988active}, moving a camera to look around occlusions~\cite{cheng2018reinforcement}, and next-best view planning~\cite{connolly1985determination}. The latter refers to approaches that select the best sensing action for specific tasks such as object instance classification~\cite{wu20153d,doumanoglou2016recovering,denzler2002information,scott2003view} and 3D reconstruction~\cite{isler2016information,kriegel2015efficient,vasquez2014volumetric,daudelin2017adaptable}. Light curtains were introduced in prior work~\cite{wang2018programmable,bartels2019agile} as an adaptive depth sensor. Prior work has also explored the use of light curtains.
\hl{\mbox{\citet{ancha20eccv}} introduced the light curtain constraint graph to compute feasible light curtains. \mbox{\citet{bartels2019agile}} were the first to empirically use random curtains to quickly discover objects in a scene. However, there are several key differences from our work. First, we solve a very different problem: while \mbox{\citet{ancha20eccv}} use light curtains to perform active bounding-box object detection in static scenes, whereas we track the safety envelope of scenes with dynamic objects. Although we build upon their constraint graph framework, we make several significant and novel contributions. Our main contribution is the safety analysis of random light curtains, which uses dynamic programming (DP) to produce theoretical guarantees on the probability of discovering objects. Providing theoretical guarantees is essential to guarantee safety, and is typically a hard task for perception systems. These works~\mbox{\cite{ancha20eccv,bartels2019agile}} do not provide any such guarantees. Additionally, we extend its constraint graph (that previously encoded only velocity constraints) to also incorporate acceleration constraints. Finally, we combine the discovery of safety envelopes using random curtains, with an ML approach that efficiently forecasts and tracks the envelope; this combination is novel, and we show that our method outperforms other approaches on this task.}
% Random curtains were first empirically used by \citet{bartels2019agile} to quickly discover objects in a scene. However, we are the first to perform a theoretical analysis and develop safety guarantees for random curtains; further, we show a novel combination of random curtains with a learning-based approach to forecast the safety envelope.

\subsection{Multi-frame depth estimation}
%\sid{Note sure how relevant this is anymore. What other related works should we look at?}
There is a large body of prior work on depth estimation across multiple frames~\cite{liu2019neural,zhang2019exploiting,cs2018depthnet,zhan2018unsupervised,matthiesdepth,wang2019recurrent,patil2020don}. \citet{liu2019neural} aggregate per-frame depth estimates across frames using Bayesian filtering. 
%However, they use both past and future frames, whereas we consider an online setting. 
\citet{matthiesdepth} use a similar Bayesian approach, but their method is only applied to controlled scenes and restricted camera motion. Other works~\cite{zhang2019exploiting,cs2018depthnet,wang2019recurrent,patil2020don} use RNNs for predicting depth maps at each frame. All of aforementioned works try to predict the full 2D depth map of the environment from monocular images. To the best of our knowledge, we are the first to use a controllable sensor to directly estimate the safety envelope of the scene.

\subsection{Safe navigation}

Many approaches for safety guaranteed navigation use 3D sensors like LiDARs~\cite{richter2018bayesian,richter2014high,schwarting2018planning} and/or cameras~\cite{richter2017safe,bajcsy2019efficient}. The sensor data is converted to occupancy grids/maps~\cite{richter2018bayesian,richter2014high,bajcsy2019efficient}; safety and collision avoidance guarantees are provided for planning under these representations. Other works use machine learning models to recognize unsafe, out-of-distribution inputs~\cite{richter2017safe} or learning to predict collision probabilites~\cite{richter2018bayesian,richter2014high}. Our work of estimating the safety envelope using a light curtain is orthogonal to these works and can leverage those methods for path planning and obstacle avoidance. 

%In contrast, we analytically derive safety guarantees and combine them with a learned model to improve efficiency. 

% to  passive perception sensors such as RGB and LiDAR for depth estimation across multiple frames. To the best of our knowledge, we are the first to use a controllable sensor and active perception algorithms for this task. \citet{liu2019neural} use RGB images for per-frame depth estimation, and aggregate these estimates across frames using Bayesian filtering. However, they use both past and future frames, whereas our method is is online and uses only past frames. \citet{matthiesdepth} use a similar Bayesian approach. However, they method is only applied to controlled scenes and restricted camera motion. Other works~\cite{zhang2019exploiting,cs2018depthnet,wang2019recurrent,patil2020don}  use RNNs for predicting depth maps at each frame. Although they perform online depth estimation, they do not perform any forecasting. In contrast, our setting requires either implicit or explicit forecasting in order to place light curtains at object surfaces in future timesteps.
\label{sec:related_work}
\section{Background on light curtains}
\label{sec:background}

\begin{figure*}
    \centering
    \subfloat[Working principle]{
        \includegraphics[width=0.23\textwidth,trim=0 0 0 0]{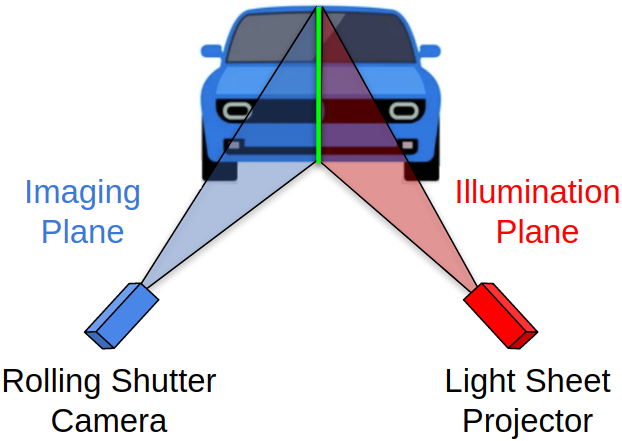}
    }
    \subfloat[Optical schematic (top view)]{
        \includegraphics[width=0.35\textwidth,trim=0 0 0 0]{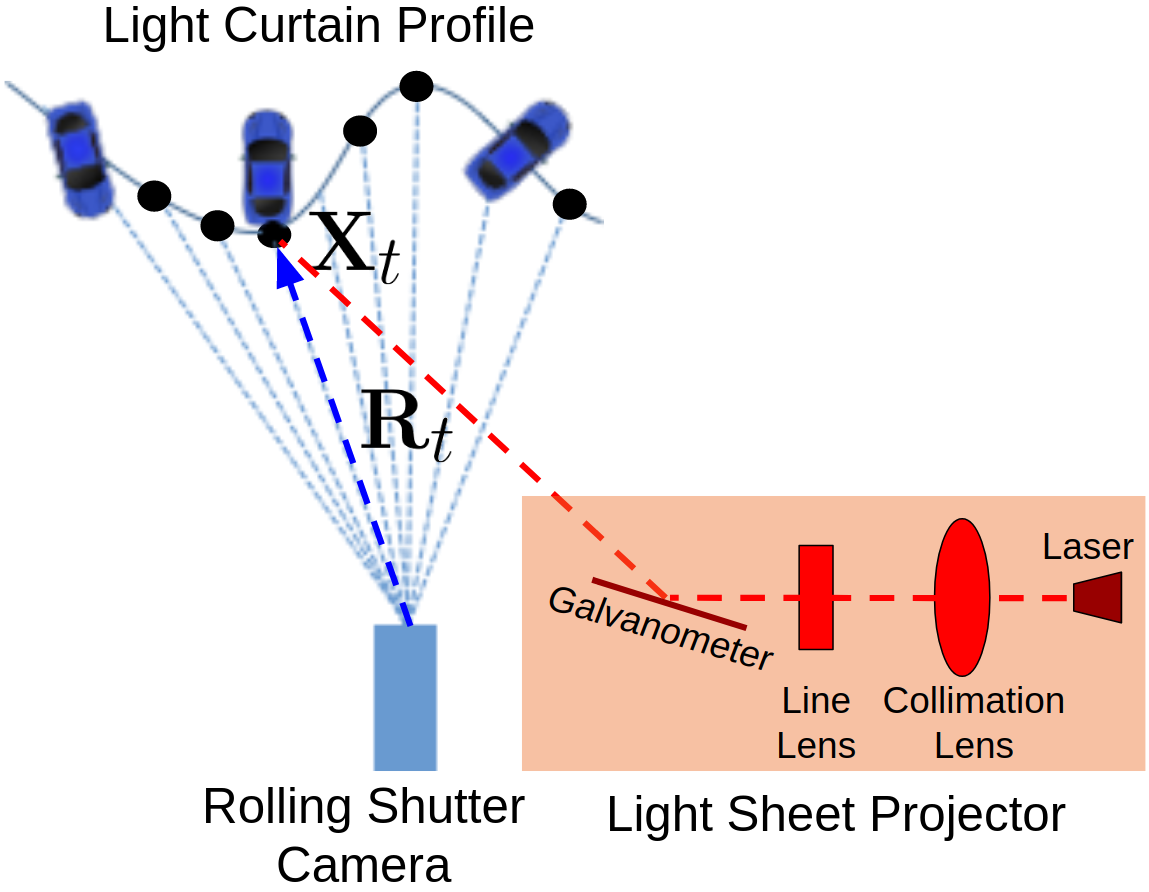}
    }
    \subfloat[Extended constraint Graph]{
        \includegraphics[width=0.35\textwidth,trim=0 0 0 0]{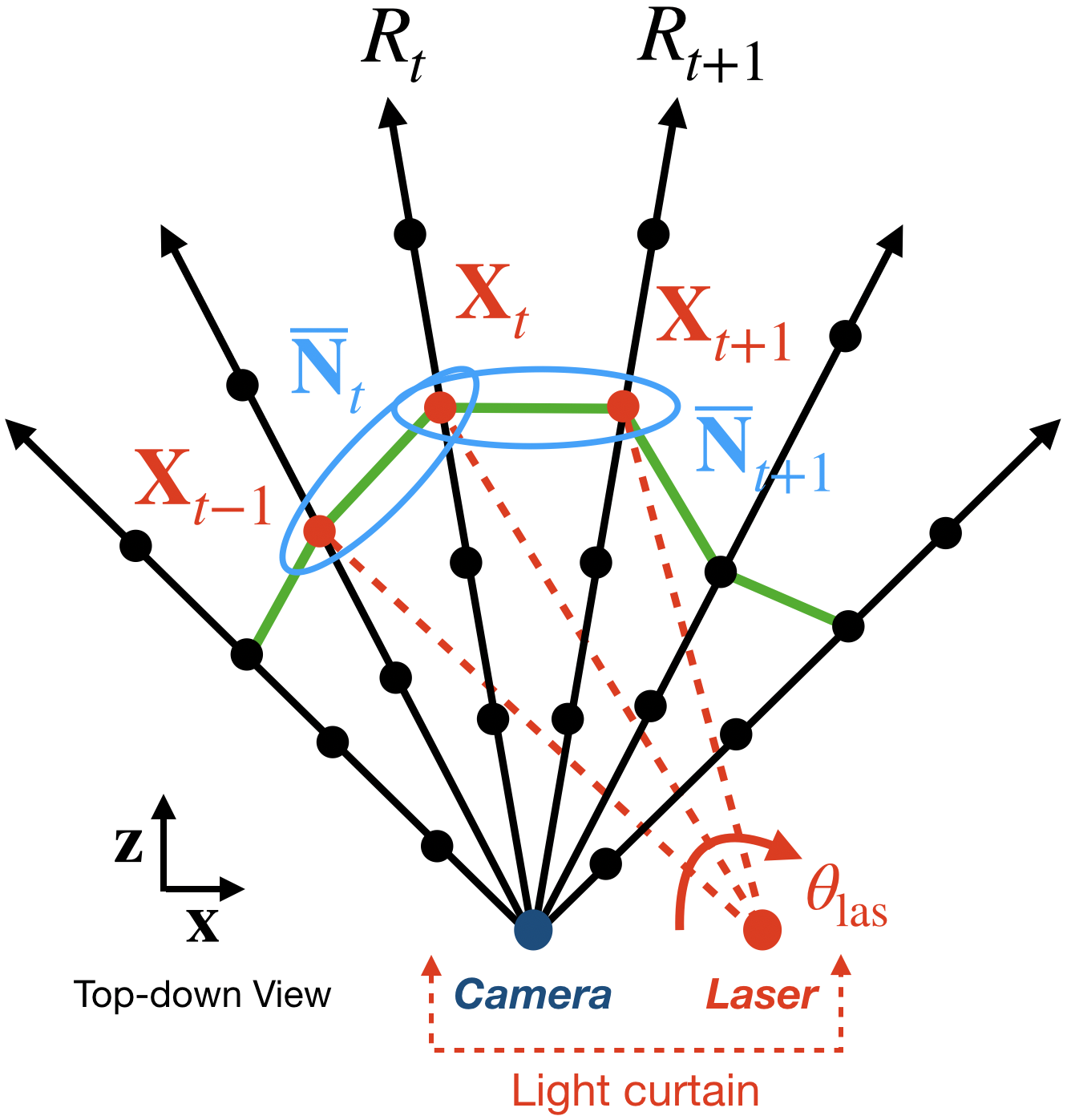}
    }
    \caption{(a, b) Illustration of programmable light curtains adapted from~\cite{ancha20eccv}. (a) An illumination plane (from the projector) and an imaging plane (of the camera) intersect to produce a light curtain. (b) A controllable galvanometer mirror rotates synchronously with a rolling shutter camera and images the points of intersection. (c) Light curtain constraint graph with our proposed extension. Black dots are control points that can be imaged, and blue ovals are extended nodes that contain two control points. Any path in this graph (shown in green) is a valid light curtain that satisfies velocity and acceleration constraints.}
    % Extended constraint graph that incorporates acceleration constraints. Each camera ray (denoted by $\R_t$) contains multiple descretized locations (denoted by $\X_t$). The locations formed the nodes of the constraint graph introduced in \citet{ancha20eccv}. We extend this graph by expanding each node (denoted by $\No_t$) to include two locations: the location on the $t$-th as well as the $(t-1)$-th camera ray. Doing so allows us to incorporate \textit{acceleration} constraints of the light curtain's galvanometer. Any feasible path (shown in green) in the extended constraint graph satisfies both velocity and acceleration constraints.}
    \label{fig:constraint-graph}
\end{figure*}

Programmable \textit{light curtains}~\cite{wang2018programmable,bartels2019agile,ancha20eccv} are a recently developed sensor for controllable depth sensing. ``Light curtains'' can be thought of as virtual surfaces placed in the environment that detect points on objects intersecting this surface. %Note: we will use the term ``light curtain'' to mean both the device as well as the virtual surface it produces, depending on the context. \dave{Maybe we can use different teminology for this?  Example: ``Light curtain sensor" vs ``Light curtain"}
%\dave{This is a good point to introduce the high level idea behind light curtains - that a light curtain is an active sensor since the user can control the position of the virtual surface to determine where to sense.  Talk about advantages and disadvantages of a light curtain compared to a traditional LiDAR device.}\sid{I have explained these points in the introduction. Not sure if we should repeat them here; this section was intended to explain the working principle behind light curtain and introduce the notation for the constraint graph.}
The working principle is illustrated in Fig.~\ref{fig:constraint-graph}(a, b). The device sweeps a vertically ruled surface by 
rotating a light sheet laser using a galvo-mirror synchronously with the vertically aligned camera's rolling shutter. Object points intersecting a vertical line of the ruled surface are imaged brightly in the corresponding camera column. We denote the top-down projection of the imaging plane corresponding to the $t$-th pixel column as a ``camera ray'' $\R_t$. The rolling shutter camera successively activates each image plane (column), corresponding to \hl{rays} $\R_1, \dots, \R_T$ from left to right, with a time difference of $\Delta t$ between successive activations. The top-down projection of the vertical line intersecting the $t$-th imaging plane lies on $\R_t$ and will be referred to as a ``control point'' $\X_t$.

\textbf{Input}: A light curtain is uniquely defined by where it intersects each camera ray $\R_t$ in the top-down view, i.e. the set of control points $(\X_1, \cdots, \X_T)$, one for each camera ray. This is the input to the light curtain device. 
%We will also refer to $(\X_1, \dots, \X_t)$ as the ``profile'' of the light curtain. 
Then, to image $\X_t$ on camera ray $\R_t$, the galvo-mirror is programmed to rotate by an angle of $\tlas(\X_t)$ that is required for the laser sheet to intersect $\R_t$ at $\X_t$. By specifying a control point $\X_t$ for each camera ray, the light curtain device can be made to image any vertically ruled surface~\cite{wang2018programmable,bartels2019agile}.

\textbf{Output}: The light curtain outputs an intensity value for each camera pixel. Since a light curtain profile is specified by a control point $\X_t$ for every camera ray $\R_t$ 
in the top-down view, we compute the maximum pixel intensity value $\I_t$ of the $t$-th pixel column and treat this as the output of the light curtain for the corresponding ray $\R_t$.

%\textbf{Physical constraints}: The rotating galvo-mirror can operate at a maximum angular velocity $\omega_\mathrm{max}$ and a maximum angular acceleration $\alpha_\mathrm{max}$. Let $\X_{t-1}, \X_t, \X_{t+1}$ be the positions imaged by the light curtain on three consecutive camera rays. These induce laser angles $\tlas(\X_{t-1}), \tlas(\X_t), \tlas(\X_{t+1})$ respectively. Since the rolling shutter camera images consecutive columns with an interval of $\Delta t$, the galvo-mirror must also rotate between the two camera rays in the same time. Denote by $\V_t = (\tlas(\X_t) - \tlas(\X_{t-1}))/\Delta t$ the angular velocity of the galvo-mirror at $\R_t$. Its angular acceleration at $\R_t$ is then $(\V_t - \V_{t-1}) / \Delta t = (\tlas(\X_{t+1}) + \tlas(\X_{t-1}) -  2 \cdot \tlas(\X_t)) / (\Delta t)^2$. Then, the light curtain velocity and acceleration constraints, in terms of the positions $\X_t$ are:
%\begin{align}
%    |\tlas(\X_t) - \tlas(\X_{t-1})| &\leq \omega_\mathrm{max} \cdot \Delta t &\forall 2 \leq t \leq T \label{eqn:vel-constraint}\\
%    |\tlas(\X_{t+1}) + \tlas(\X_{t-1}) -  2 \tlas(\X_t)| &\leq \alpha_\mathrm{max} \cdot (\Delta t)^2 &\forall 3 \leq t \leq T \label{eqn:acc-constraint}
%\end{align}
%\section{Extended light curtain constraint graph}
\section{Generating Feasible Light Curtains}
\label{sec:extended-graph}
The set of realizable curtains depend on the physical constraints imposed by the real device.
The rotating galvo-mirror can operate at a maximum angular velocity $\omega_\mathrm{max}$ and a maximum angular acceleration $\alpha_\mathrm{max}$. Let $\X_{t-1}, \X_t, \X_{t+1}$ be the control points imaged by the light curtain on three consecutive camera rays. These induce laser angles $\tlas(\X_{t-1}), \tlas(\X_t), \tlas(\X_{t+1})$ respectively. 
%Since the camera images consecutive columns with an interval of $\Delta t$, the galvo-mirror must also rotate between the two camera rays in the same time. 
Let $\V_t = (\tlas(\X_t) - \tlas(\X_{t-1}))/\Delta t$ be the angular velocity of the galvo-mirror at $\R_t$. Its angular acceleration at $\R_t$ is $(\V_{t+1} - \V_t) / \Delta t = (\tlas(\X_{t+1}) + \tlas(\X_{t-1}) -  2 \cdot \tlas(\X_t)) / (\Delta t)^2$. Then, the light curtain velocity and acceleration constraints, in terms of the control points $\X_t$, are:
\begin{align}
   |\tlas(\X_t) - \tlas(\X_{t-1})| &\leq \omega_\mathrm{max} \cdot \Delta t & 2 \leq t \leq T \label{eqn:vel-constraint}\\
    |\tlas(\X_{t+1}) + \tlas(\X_{t-1}) -  2 \tlas(\X_t)| &\leq \alpha_\mathrm{max} \cdot (\Delta t)^2 & 2 \leq t < T \label{eqn:acc-constraint}
\end{align}
In order to compute feasible light curtains that satisfy physical constraints, \citet{ancha20eccv} introduced a ``light curtain constraint graph,'' denoted by $\Go$. $\Go$ has two components: a set of nodes $\No_t$ associated with each camera ray and edges between nodes $\No_t$ and $\No_{t+1}$ on consecutive camera rays. Nodes are designed to store information that fully captures the \textit{state} of the galvo-mirror when imaging the ray $\R_t$. An edge exists from $\No_t$ to $\No_{t+1}$ \textit{iff} the galvo-mirror is able to transition from the state defined by $\No_t$ to the state defined by $\No_{t+1}$ without violating any of velocity constraints (Eqn.~(\ref{eqn:vel-constraint})). 

%Thus, the constraint graph defines the set of all feasible light curtains. 

%Since each node represents one camera ray, the adjacent nodes could be tested to not violate the velocity constraint (Eqn.~(\ref{eqn:vel-constraint})). 

\hl{\mbox{\citet{ancha20eccv}} defined the state to be $\No_t = \X_t$ (i.e. contain only one control point).}
But this representation does not ensure that acceleration constraints of Eqn.~\ref{eqn:acc-constraint}, that depend on three consecutive control points, are satisfied. This can produce light curtain profiles which require the galvo-mirror to change its angular velocity more abruptly than its physical torque limits can allow, \hl{resulting in hardware errors}. Thus, we extend the definition of a node $\No_t$ at $t$ to store control points on the current ray \textit{and} the previous ray, as $\Ne_t = (\X_{t-1}, \X_t)$ (see Fig.~\ref{fig:constraint-graph}~(c)). Intuitively, $\Ne$ contains information about the angular position and velocity of the galvo-mirror. This allows us to incorporate acceleration constraints by creating an edge between nodes $(\X_{t-1}, \X_t)$ and $(\X_t, \X_{t+1})$, \textit{iff} $\X_{t-1}, \X_t, \X_{t+1}$ satisfy the velocity and acceleration constraints defined in Equations~(\ref{eqn:vel-constraint},~\ref{eqn:acc-constraint}). Thus, any path in the extended graph $\Ge$ represents a feasible light curtain that satisfies both constraints. The acceleration constraints also serve to limit the increase in the number of nodes with feasible edges, keeping the graph size manageable. 

\section{Random curtains \& theoretical guarantees}
\label{sec:random-curtain-analysis}

Recall that the light curtain will only sense the parts of the scene where the curtain is placed. Thus we must decide where to place the curtain in order to sense the scene and estimate the safety envelope.
Our proposed method uses a combination of random curtains as well as learned forecasting to estimate the safety envelope of an unknown scene.
In this section, we show how a random curtain can be sampled from the extended constraint graph $\Ge$ and how to analytically compute the probability of a random curtain detecting an obstacle, which helps to probabilistically guarantee obstacle detection of our overall method.

\subsection{Sampling random curtains from the constraint graph}
\label{sec:random-curtain-analysis:sampling-random-curtain}

% \dave{You should first motivate why you would want to sample a random light curtain}
% A random light curtain is a light curtain whose placement is randomly generated (see Figure~\sid{refer to a figure that shows a random curtain}). A single random curtain can image a large number of random positions in the scene. In Section~\ref{sec:experiments}, we show that random curtains can quickly explore the scene and detect untracked objects with a large probability, leading to significant improvements in depth tracking performance. Furthermore, in Section~\ref{sec:approach:theoretical-analysis} we develop an approach that provides theoretical guarantees for the probability of detection, that can be used to certify the safety and robustness of our system.

%We will now show how a random curtain can be sampled from the extended constraint graph $\Ge$. %This ensures that the generated curtain will respect the light curtain constraints.
%We use random curtains to discover the safety envelope of an unknown scene. 
% The extended constraint graph described in Section~\ref{sec:extended-graph} provides a framework to do so. Any path through the extended constraint graph represents a valid light curtain that satisfies both velocity and acceleration constraints of the galvanometer.
First, we need to define a probability distribution over the set of valid curtains in order to sample from it. We do so by defining, for each node $\Ne_t=(\X_{t-1}, \X_t)$, a \textit{transition probability distribution} $P(\X_{t+1} \mid \X_{t-1}, \X_t)$. This denotes the probability of transitioning from imaging the control points $\X_{t-1}, \X_t$ on the previous and current camera rays to the control point $\X_{t+1}$ on the next ray. We constrain $P(\X_{t+1} \mid \X_{t-1}, \X_t)$ to equal $0$ if there is no edge from node $(\X_{t-1}, \X_t)$ to node $(\X_t, \X_{t+1})$; an edge will exist \textit{iff} the transition $\X_{t-1} \rightarrow \X_t \rightarrow \X_{t+1}$ satisfies the light curtain constraints. Thus, $P(\X_{t+1} \mid \X_{t-1}, \X_t)$ defines a probability distribution over the  neighbors %$(\X_t, \X_{t+1})$ 
of $\Ne_t$ in the constraint graph.
% also be thought of as defining a probability distribution over the Note that for any node $\Ne_{t+1} \not\in \nbhood(\Ne_t)$ not belonging to the forward neighborhood of $\Ne_t$, $\ptrans(\Ne_{t+1} \mid \Ne_t) = 0$, since we only want to transition along edges. In the simplest case, $\ptrans(\Ne_{t+1} \mid \Ne_t)$ will be uniform over $\nbhood(\Ne_t)$.\\

The transition probability distribution enables an algorithm to sequentially generate a random curtain. We begin by sampling the control points $\Ne_2 = (\X_1, \X_2)$ according to an initial probability distribution $P(\Ne_2)$.
% At the $(t-1)$-th iteration, we will have sampled the first $t$ positions $(\X_1, \dots, \X_t)$ of the random curtain. 
At the $t$-th iteration, we sample $\X_{t+1}$ according to the transition probability distribution $\X_{t+1} \sim P(\X_{t+1} \mid \X_{t-1}, \X_t)$ and add $\X_{t+1}$ to the current set of sampled control points. After $(T-1)$ iterations, this process generates a full random curtain. Pseudo-code for this process is found in Algorithm 1 in Appendix~\ref{sec:appendix:sampling-curtains}. Our random curtain sampling process provides the flexibility to design any initial and transition probability distribution. See Appendix~\ref{sec:appendix:sampling-curtains} for a discussion on various choices of the transition probability distribution, where we also provide a theoretical and empirical justification to use one distribution in particular.

\subsection{Theoretical guarantees for random curtains} 
\label{sec:random-curtain-analysis:theoretical-analysis}

In this section, we first describe a procedure to detect objects in a scene using the output of a random light curtain placement. Then, we develop a method that runs dynamic programming on $\Ge$ to analytically
%\sid {I'm looking for words to convey that probability estimates using DP are accurate as opposed to noisy/stochastic estimates: exacty/precisely/accurately/deterministically} 
compute a random curtain's probability of detecting a specific object. This provides probabilistic safety guarantees on how well a random curtain can discover the safety envelope of an object.

\textbf{Detection using light curtains:} Consider an object in the scene whose visible surface intersects each camera ray at the positions $O_{1:T}$.
%at every camera ray $\R_t$. 
This representation captures the position, shape and size of the object from the top-down view. Let $\{\X_t\}_{t=1}^T$ be the set of control points for a light curtain placed in the scene. The light curtain will produce an intensity $\I_t(\X_t, O_t)$ at each control point $\X_t$ that is sampled by the light curtain device. Note that $\I_t$ is a function of the position of the object as well as the position of the light curtain; the intensity increases as the distance between $\X_t$ and $O_t$ reduces and is the highest when $\X_t$ and $O_t$ coincide. We say that an object has been detected at control point $\X_t$ if the intensity $\I_t$ is above a threshold $\tau$; the intensity threshold is used to account for noise in the image. We define a binary detection variable to indicate whether a detection of an object occurred at position $O_t$ at control point $\X_t$ as $\dd_t(\X_t, O_t) = [\I_t(\X_t, O_t) > \tau]$, where $[\cdot]$ is the indicator function. % (Iversion bracket notation).
%, which in Iversion bracket notation means that $\dd_t$ is 1 \textit{iff} $\I_t$ is greater than $\tau$, and 0 otherwise. 
We declare that an object has been detected by a light curtain if it is detected on any of its control points. 
Formally, we define a binary detection variable to indicate whether a detection of object $O_{1:T}$ occurred at any of its control point $\X_{1:T}$ as  $\dd(\X_{1:T}, O_{1:T}) = \bigvee_{t=1}^T \dd_t(\X_t, O_t)$, (where $\bigvee$ is `logical or' operator).
%: $\dd = 1$ \textit{iff} $\exists t: \dd_t = 1$).
% \textbf{Detection probability of random curtains:} In the remainder of this section, we assume that light curtains $\X_{1:T}$ are sampled using the constraint graph $\Ge$ as described in Sec.~\ref{sec:random-curtain-analysis:sampling-random-curtain}, and that an object of known positions $O_{1:T}$ exists in the scene. Note that $\X_{1:T}$ is now a random variable, and so are the variables $\I_t, \dd_t$ and  $\dd$ that are functions of $\X_{1:T}$. Some random curtains sampled from $\Ge$ might detect the object, while others might not. 
Our objective is then to compute the \textit{detection probability}, denoted as $P( \dd(\X_{1:T}, O_{1:T}))$, which is the probability that a curtain sampled from $\Ge$ (using the sampling procedure described in Sec.~\ref{sec:random-curtain-analysis:sampling-random-curtain}) will detect the object $O_{1:T}$; below we will use the simpler notation $P(\dd)$ to denote the detection probability.

\textbf{Theoretical guarantees using dynamic programming:} The simplest method to compute the detection probability for a given object is to sample a large number of random curtains and output the average number of curtains that detect the object. However, a large number of samples would be needed to provide accurate estimates of the detection probability; further, this procedure is stochastic and the probability estimate will only be approximate. Instead, we propose utilizing the known structure of the constraint graph and the transition probabilities; we will apply dynamic programming to compute the detection probability both efficiently and analytically. 

%We perform a comparison between our approach and the sampling based approach in Section~\ref{sec:experiments:analysis} to show that the two methods give the same results, even though ours is much more efficient.

Our analytic method for computing the detection probability proceeds as follows:
we first compute the value of the detection event $\dd_t(\X_t, O_t)$ at every possible control point $\X_t$ that is part of a node in the constraint graph
%<=\dave - how about this? <---$
% based on what the reader has read so far, it would seems that nodes, control pts etc. are continuous. How could nodes in a graph be continuous?
% X_t are defined to be the intersections of the camera ray and the laser beam. When we say that (X_1, ..., X_T) represents any curtain, it implies that they are continuous,S old 
% Should we jump on a quick zoom call? Yes We can use the same link as this mornign
%\dave{What is $\set_t$?  This has not been defined} \sid{this was defined previously in Sec. 4 when I introduced the constraint graph, but was later removed. We need to mention somewhere that the constraint graph is discrete (and $\set_t$ is the set of discrete control points) and at every $t$ in the constraint graph 
$\Ge$ i.e. we compute whether or not a curtain placed at $\X_t$ is able to detect the object. 
%This is possible because that all physical properties of the light curtain device are known; these include the intrinsics of the camera and the power, thickness and divergence of the laser beam. 
Given the positions $O_{1:T}$ of an object, as well as the physical properties of the light curtain device (intrinsics of the camera and the power, thickness and divergence of the laser beam), we use a light curtain simulator to compute $\I_t(\X_t, O_t)$ using standard raytracing and rendering, for any arbitrary control point $\X_t$.
% Hence, we can compute the intensity and $\dd(\X_t=\x{t}{i}, O_t)$ for every position $\x{t}{i}$ in $\Ge$.
%Henceforth, we will drop the conditioning on $O_t$ for simplicity.

% \dave{Remove lowercase variables from this paragraph} Our method works as follows. To compute the detection probability $P(\dd)$, we first decompose the overall problem into simpler sub-problems by defining the \textit{sub-curtain detection probability} $\pdet(\x{t-1}{i}, \x{t}{j}) = P(\bigvee_{t'=t}^T \dd_t \mid \X_{t-1}=\x{t-1}{i}, \X_t=\x{t}{j})$. This is the probability that a random sub-curtain starting on the node $(\X_{t-1}=\x{t-1}{i}, \X_t=\x{t}{j})$ and ending on the last camera ray $\R_T$, detects the object $O$ at some point between rays $\R_t$ and $\R_T$. Note that the overall detection probability can be written in terms of the sub-curtain detection probabilities of the second ray (the first node in the graph) as $P(\dd) = \sum_{\set_2} P(\X_1 = \x{1}{i}, \X_2 = \x{2}{j}) \cdot \pdet(\X_1 = \x{1}{i}, \X_2 = \x{2}{j})$ \dave{Is this sum supposed to be over the values of $X_1$, $X_2$?}. This is the sum of the detection probabilities of a random curtain starting from the initial nodes, weighted by the probability of the nodes being sampled from the initial distribution. Conveniently, the sub-curtain detection probabilities satisfy a simple recursive equation:

 To compute the detection probability $P(\dd)$, we first define the notion of a ``sub-curtain," which is a subset of the control points $\X_{t:T}$ which start at ray $R_t$ and ends on ray $R_T$.
 We can decompose the overall problem of computing $P(\dd)$ into simpler sub-problems by defining the \textit{sub-curtain detection probability} $\pdet(\X_{t-1}, \X_{t})$. This is the probability that any random sub-curtain starting at $(\X_{t-1}, \X_t)$ and ending on the last camera ray $\R_T$ detects the object $O$ at some point between rays $\R_t$ and $\R_T$. Using this definition, we can write the sub-curtain detection probability as \hl{$\pdet(\X_{t-1}, \X_{t}) = P(\bigvee_{t'=t}^T \dd_{t'}(\X_{t'}, O_{t'}) \mid \X_{t-1}, \X_t)$}.
 
 Note that the overall curtain detection probability can be written in terms of the sub-curtain detection probabilities of the second ray (the first set of nodes in the graph) as 
 \begin{align}
     P(\dd) = \sum_{\X_1, \X_2} \pdet(\X_1, \X_2)\ P(\X_1, \X_2).
     \label{eq:overall_det_prob}
 \end{align} This is the sum of the detection probabilities of a random curtain starting from the initial nodes $\pdet(\X_1, \X_2)$, weighted by the probability of the nodes being sampled from the initial distribution $P(\X_1, \X_2)$. Conveniently, the sub-curtain detection probabilities satisfy a simple recursive equation:
\begin{align}
  &\pdet(\X_{t-1}, \X_{t}) = \nonumber\\
  &\ \ \ \ \begin{cases}
      1 \hspace{14em} \text{if\ } \dd_t(\X_t, O_t) = 1\\ \\
      \sum\limits_{\X_{t+1}} \pdet(\X_{t}, \X_{t+1})\ P(\X_{t+1} \mid \X_{t-1}, \X_t ) \ \ \text{otherwise}
  \end{cases}
  \label{eq:recursive}
\end{align}
Intuitively, if the control point $\X_t$ is able to detect the object, then the sub-curtain detection probability is 1 regardless of how the sub-curtain is placed on the later rays. If not, then the detection probability should be equal to the sum of the sub-curtain detection probabilities of the nodes the curtain transitions to, weighted by the transition probabilities.

This recursive relationship can be exploited by successively computing the sub-curtain detection probabilities from the last ray to the first. The sub-curtain detection probabilities on the last ray will simply be either $1$ or $0$, based on whether the object is detected there or not. After having computed sub-curtain detection probabilities for all rays between $t+1$ and $T$, the probabilities for nodes at ray $t$ can be computed using the above recursive formula (Eqn.~\ref{eq:recursive}). Finally, after obtaining the probabilities for nodes on the second ray, the overall curtain detection probability can be computed as described using Eqn.~\ref{eq:overall_det_prob}. Pseudocode for this method can be found in Algorithm 2 in Appendix~\ref{sec:appendix:dp}. \hl{A discussion on the computational complexity of the extended constraint graph $\Ge$ (which contains more nodes and edges than $\Go$) can be found in Appendix~\mbox{\ref{sec:appendix:runtime}}}.

We have created a web-based demo \hl{(available on the project website)} that computes the probability of a random curtain detecting an object with a user-specified shape in the top-down view. It also performs analysis of the detection probability as a function of the number of light curtains placed.
In Section~\ref{sec:experiments:analysis}, we use this method to analyze the detection probability of random curtains as a function of the object size and number of curtain placements. We compare it against a sampling-based approach and show that our method gives the same results but with an efficient analytical computation.
\section{Learning to forecast safety envelopes}
\label{sec:learning-forecasting}

Random curtains can help discover the safety envelope of unknown objects in a scene. However, once a part of the envelope is discovered, an efficient way to estimate the safety envelope in future timesteps is to \textit{forecast} how the envelope will move with time and \textit{track} the envelope by placing a new light curtain at the forecasted locations. In this section, we describe how to train a deep neural network to forecast safety envelopes and combine them with random curtains.

\textbf{Problem setup:} 
We call any algorithm that attempts to forecast the safety envelope as a ``forecasting policy''. 
%\sid{should I use the term ``policy", or just ``model"?}. 
We assume that a forecasting policy is provided with the ground truth safety envelope of the scene in the first timestep. In the real world, this can be done by running one of the less efficient baseline methods once when the light curtain is first started, until the light curtain is initialized.  After the initialization, the learning-based method is used for more efficient light curtain tracking.
% This mimics the common scenario where a robot is initialized in a familiar environment (but that may evolve arbitrarily), such as an autonomous vehicle starting a trip from its garage, or a delivery robot initialized at a storage depot. 
The policy always has access to all previous light curtain measurements. At every timestep, the policy is required to forecast the location of the safety envelope for the next timestep. Then, the next light curtain will be placed at the forecasted location. \hl{To leverage the benefits of random curtains that can discover unknown objects, we place random light curtains while the forecasting method predicts the safety envelope of the next timestep}. We allow random curtains to override the forecasted curtain: if the random curtain obtains an intensity $\I_t$ on camera ray $\R_t$ that is above a threshold $\tau$, the control point of the forecasted curtain for ray $\R_t$ is immediately updated to that of the random curtain, i.e. the random curtain overrides the forecasted curtain if the random curtain detects an object. \hl{See Appendix~\mbox{\ref{sec:appendix:pipeline}} for details of our efficient, parallelized implementation of random curtain placement and forecasting, as well as an analysis of the pipeline's runtime.}

%\textbf{Training procedure:} 

%\sid{is there a better word than `handcrafted'?} 
\textbf{Handcrafted policy: }
First, we define a simple, hand-specified light curtain placement policy; this policy will serve both as a baseline and as an input to our neural network, described below. The policy conservatively forecasts a fixed decrease in the depth of the safety envelope for ray $\R_t$ if the ray's intensity $\I_t$ is above a threshold (indicative of the presence of an object), and forecasts a fixed increase in depth otherwise. By alternating between increasing and decreasing the depth of the forecasted curtain, this policy can roughly track the safety envelope. However, since the forecasted changes in depth are hand-defined, it is not designed to handle large object motions, nor will it accurately converge to the correct depth for stationary objects. %Regardless, we found that it works reasonably well in practice.

\textbf{Neural network forecasting policy:} We use a 2D convolutional neural network to forecast safety envelopes in the next timestep. It takes as input (1) the intensities $\I_t$ returned by previous $k$ light curtain placements, (2) the positions of the previous $k$ light curtain placements, and (3) the outputs of the handcrafted policy described above (\hl{this helps avoid local minima during training and provides useful information to the network}). For more details about the architecture of our network, please see Appendix~\ref{sec:appendix:architectures}.

We assume access to ground truth safety envelopes at training time. This can be directly obtained in simulated environments or from auxiliary sensors such as LiDAR in the real world. Because a light curtain is an active sensor, the data that it collects depends on the forecasting policy. Thus to train our network, we use DAgger~\cite{ross2011reduction}, a widely-used imitation learning algorithm to train a policy with expert or ground-truth supervision across multiple timesteps. We use the Huber loss~\cite{huber1992robust} between the predicted and ground truth safety envelopes as our training loss. The Huber loss is designed to produce stable gradients while being robust to outliers.

\section{Experiments}
\label{sec:experiments}

\begin{figure}
  \centering
  \includegraphics[scale=0.4,trim=0 0 0 0]{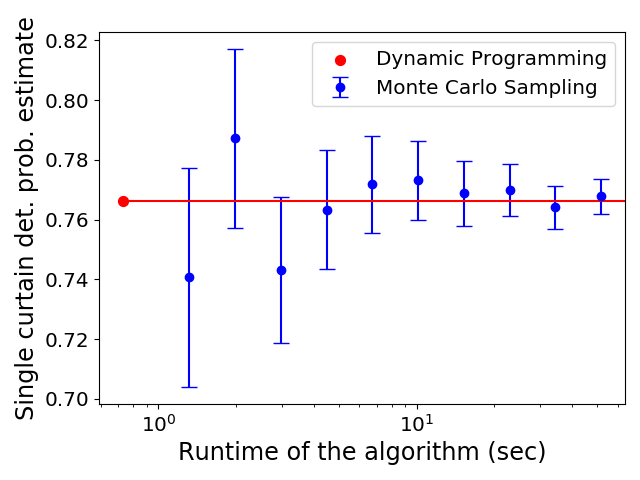}
  \caption{Comparison of our dynamic programming approach with a Monte Carlo sampling based approach to estimate the probability of a random curtain detecting an object of size $2m \times 2m$. The $x$-axis shows the runtime of the methods (in seconds), and the $y$-axis shows estimated detection probability. Our method (in red) is both \textit{exact} and fast; it quickly produces a single and precise point estimate. Monte Carlo sampling (in blue) samples a large number of random curtain and returns the average number of curtains that intersect the object. The estimate is stochastic and the $95\%$-confidence intervals are shown in blue. While the Monte Carlo estimate eventually converges to the true probability, it is inherently noisy and orders of magnitude slower than our method.}
  \label{fig:analysis:monte-carlo}
\end{figure}

\begin{figure*}
  \centering
  (a) \includegraphics[scale=.4,trim=0 0 0 0]{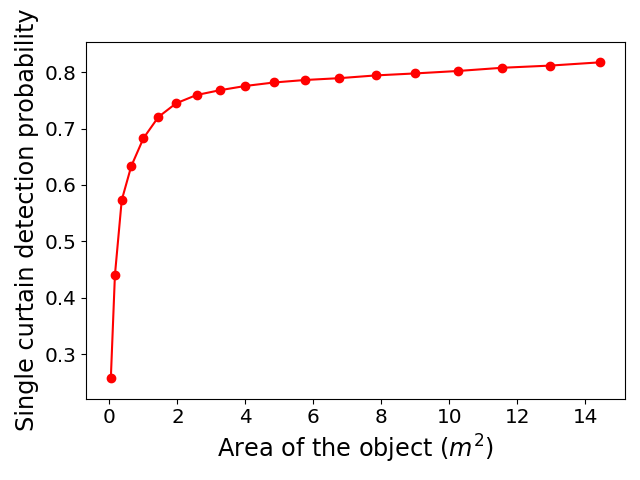}
  (b) \includegraphics[scale=.4,trim=0 0 0 0]{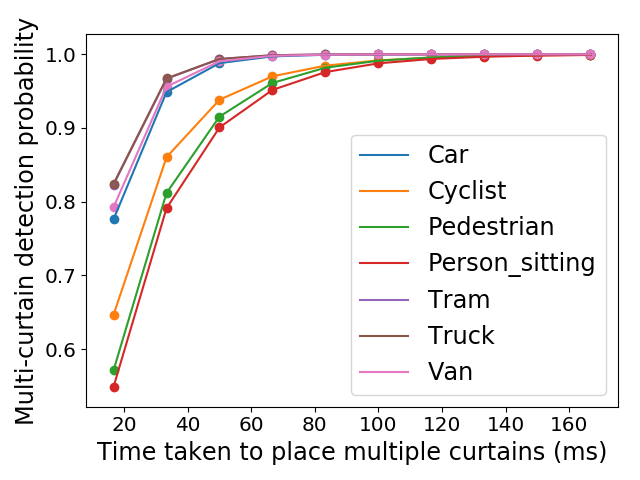}
  \caption{(a) The probability of random curtains detecting objects of various areas. For an object of a fixed area, we average the probability across various orientations of the object. Larger objects are detected with higher probability. (b) We show the detection probability of canonical objects from classes in the KITTI~\cite{Geiger2012CVPR} dataset. For each object class, we construct a ``canonical object" by
  averaging the dimensions of all labeled instances of that class in the KITTI dataset. Larger object classes are detected with a higher probability, as expected. We also show the detection probability as a function of the number of light curtains placed. The detection probability increases exponentially with the number of light curtain placements.}
  \label{fig:analysis:area-and-curtains}
\end{figure*}

\begin{figure*}
  \centering
  \newcommand{\scale}{1.0}
  \includegraphics[width=\scale\textwidth]{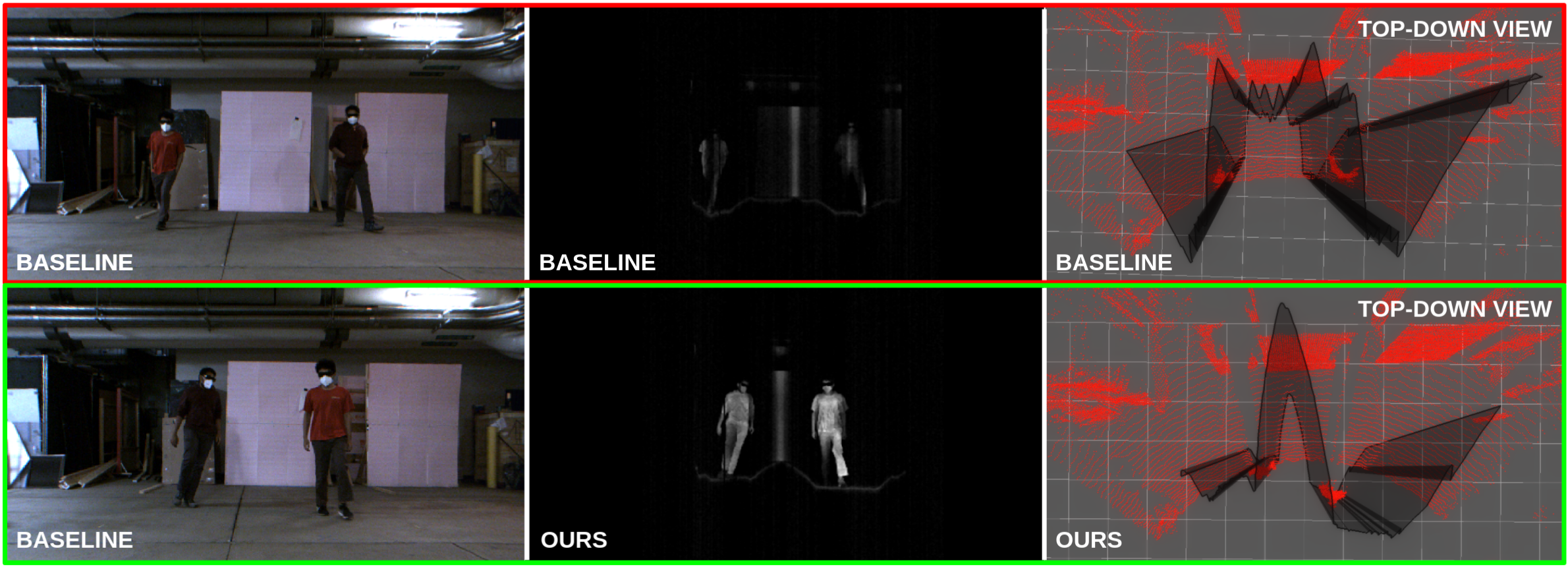}
  \caption{Qualitative results in a real-world environment with two walking pedestrians, comparing a hand-crafted baseline policy (top-row) with our method (bottom-row). \textit{Left column:} contains RGB scene images. \textit{Middle column:} contains the light curtain images, where higher intensity means a closer intersection between the light curtain and the object surfaces (i.e. a better estimation of the safety envelope). Since our method learns to forecast the safety envelope, it estimates the envelope more accurately and produces higher light curtain intensity. \textit{Right column} (top-down view): the black surfaces are the estimated safety envelopes, and red points show a LiDAR point cloud (only used to aid visualization). Our forecasting method's estimate of the safety envelope hugs the pedestrians more tightly and looks smoother. The hand-crafted baseline works by continuously moving the curtain back and forth, creating a jagged profile and preventing it from enveloping objects tightly.}
  \label{fig:real_world_experiment}
\end{figure*}

\begin{figure*}
    \centering
    \includegraphics[width=1.0\textwidth]{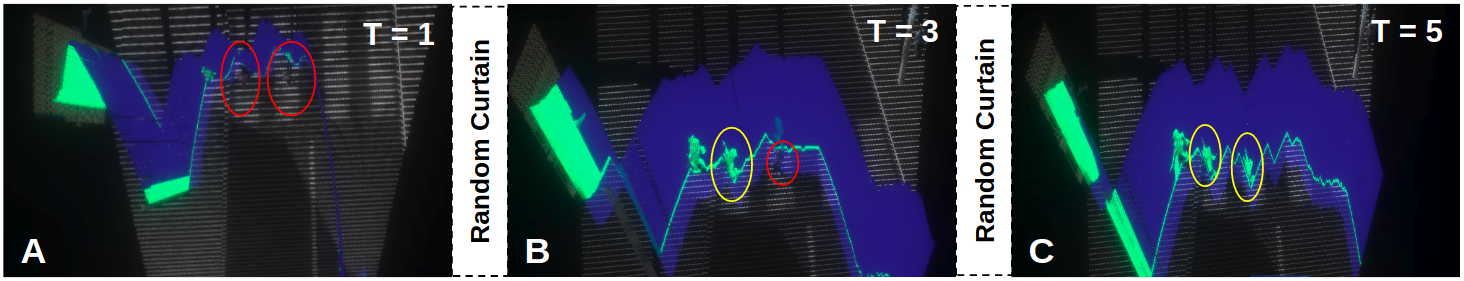}
    \caption{We illustrate the benefits of placing random curtains (that come with probabilistic guarantees of obstacle detection) while estimating safety envelopes, shown in SYNTHIA~\cite{synthia}, a simulated urban driving environment. The blue surfaces are the estimated safety envelopes, and the green points show regions of high light curtain intensity (higher intensity corresponds to better estimation). There are three pedestrians in the scene. (a) Our forecasting model fails to locate two pedestrians (red circles). (b) The first random curtain leads to the discovery of one pedestrian (yellow). (c) The second random curtain helps discover the other pedestrian (second yellow circle). The safety envelope of all pedestrians has now been detected.}
    \label{fig:random_curtain_detects}
\end{figure*}

\begin{figure*}
    \centering
    \includegraphics[width=1.0\textwidth]{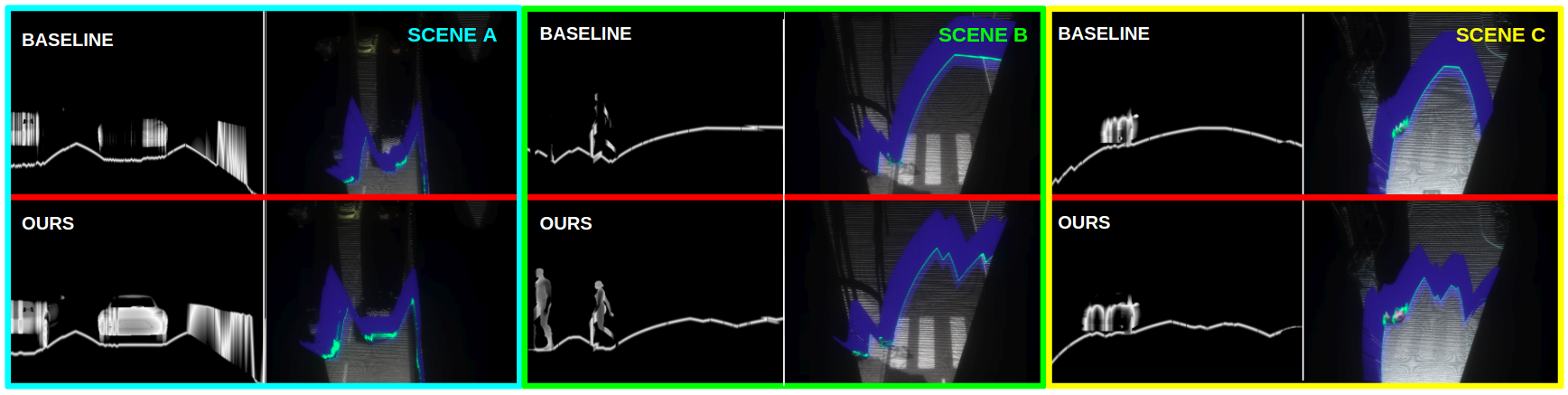}
    \caption{Comparison of the safety envelope estimation between a hand-crafted baseline policy (top row) and a trained neural network (bottom row), in three simulated urban driving scenes from the SYNTHIA~\cite{synthia} dataset. For each scene and method, the left column shows the intensity image of the light curtain; higher intensities correspond to closer intersection of the light curtain and object surfaces, implying better estimation of the safety envelope. The right column shows the light curtain profile in the scene. The trained network estimates the safety envelopes more accurately than the handcrafted baseline policy.}
    \label{fig:qualitative_synthia}
\end{figure*}

\begin{table*}
  \centering
  \begin{tabular}{?c?c?c|c|c|c|c|c|c|c|c?} 
   \Xhline{0.8pt}
    & Huber loss & \makecell{RMSE\\Linear} & \makecell{RMSE\\Log} & \makecell{RMSE\\Log Scale-Inv.} & \makecell{Absolute\\Relative Diff.} & \makecell{Squared\\Relative Diff.} & \makecell{Thresh\\($1.25$)} & \makecell{Thresh\\($1.25^2$)} & \makecell{Thresh\\($1.25^3$)}\\
     \cline{2-10}
    & $\downarrow$ &  $\downarrow$ & $\downarrow$ & $\downarrow$ & $\downarrow$ & $\downarrow$ & $\uparrow$ & $\uparrow$ & $\uparrow$\\
    % \Xhline{0.8pt}
    %  Handcrafted baseline & \xmark & 0.1989 & 2.5811 & 0.2040 & 0.0904 & 0.2162 & 2.0308 & 0.6321 & 0.7321 & 0.7657\\
    %  \hline
    % 1D-CNN & \xmark & 0.1522 &  2.3856 & 0.2176 & 0.1076 & 0.1750 & 0.9482 & 0.5842 & 0.7197 & \textbf{0.7868}\\
    % \hline
    % 1D-GNN & \xmark & 0.1584 & 2.2114 & 0.1835 & \textbf{0.0839} & 0.1772 & 1.1999 & 0.6546 & 0.7381 & 0.7710\\
    % \hline
    % Ours w/o Forecasting & \xmark & 0.1691 & 2.6047 & 0.2288 & 0.1158 & 0.1927 & 1.2555 & 0.6109 & 0.7114 & 0.7654\\
    % \hline
    % Ours w/o Baseline input & \xmark & 0.1556 & 2.5987 & 0.2273 & 0.1135 & 0.1797 & 1.1063 & 0.6021 & 0.7094 & 0.7683\\
    % \hline
    % % Ours w/o 5 frames & \xmark & & & & & & & & &\\
    % % \hline
    % \textbf{Ours} & \xmark & \textbf{0.1220} & \textbf{2.0332} & \textbf{0.1724} & 0.0888 &  \textbf{0.1411} & \textbf{0.9070} & \textbf{0.6752} & \textbf{0.7450} & 0.7852\\
   \Xhline{0.8pt}
   Handcrafted baseline & 0.1145 & 1.9279 & 0.1522 & 0.0721 & 0.1345 & 1.0731 & 0.6847 & 0.7765 & 0.8022\\
    \hline
    Random curtain only & 0.1484 & 2.2708 & 0.1953 & 0.0852 & 0.1698 & 1.2280 & 0.6066 & 0.7392 & 0.7860\\
     \hline
    1D-CNN & 0.0896 & 1.7124 & 0.1372 & 0.0731 & 0.1101 & 0.7076 & 0.7159 & 0.7900 & 0.8138\\
    \hline
    1D-GNN & 0.1074 & 1.6763 & 0.1377 & 0.0669 & 0.1256 & 0.8916 & 0.7081 & 0.7827 & 0.8037\\
    \hline
    Ours w/o Random curtains & 0.1220 & 2.0332 & 0.1724 & 0.0888 & 0.1411 & 0.9070 & 0.6752 & 0.7450 & 0.7852\\
    \hline
    Ours w/o Forecasting & 0.0960 & 1.7495 & 0.1428 & 0.0741 & 0.1163 & 0.6815 & 0.7010 & 0.7742 & 0.8024\\
    \hline
    Ours w/o Baseline input & 0.0949 & 1.8569 & 0.1600 & 0.0910 & 0.1148 & 0.7315 & 0.7082 & 0.7740 & 0.7967\\
    \hline
    % Ours w/o 5 frames & \cmark & & & & & & & & &\\
    % \hline
    \textbf{Ours} & \textbf{0.0567} & \textbf{1.4574} & \textbf{0.1146} & \textbf{0.0655} & \textbf{0.0760} & \textbf{0.3662} & \textbf{0.7419} & \textbf{0.8035} & \textbf{0.8211}\\
   \Xhline{0.8pt}
  \end{tabular}
  \caption{Performance of safety envelope estimation on the SYNTHIA~\cite{synthia} urban driving dataset under various metrics.}
  \label{table:synthia}
\end{table*}

\begin{table*}
  \centering
  \begin{tabular}{?c?c?c|c|c|c|c|c|c|c|c?} 
   \Xhline{0.8pt}
    & & Huber loss & \makecell{RMSE\\Linear} & \makecell{RMSE\\Log} & \makecell{RMSE\\Log Scale-Inv.} & \makecell{Absolute\\Relative Diff.} & \makecell{Squared\\Relative Diff.} & \makecell{Thresh\\($1.25$)} & \makecell{Thresh\\($1.25^2$)} & \makecell{Thresh\\($1.25^3$)}\\
    \cline{3-11}
    & & $\downarrow$ &  $\downarrow$ & $\downarrow$ & $\downarrow$ & $\downarrow$ & $\downarrow$ & $\uparrow$ & $\uparrow$ & $\uparrow$\\
   \Xhline{0.8pt}
   \multirow{2}{*}{\makecell{\textit{Slow}\\\textit{Walking}}} & Handcrafted baseline & 0.0997 & 0.9908 & 0.1881 & \textbf{0.1015} & 0.1371 & 0.2267 & 0.8336 & \textbf{0.9369} &
   \textbf{0.9760}\\
   \cline{2-11}
   & \textbf{Ours} & \textbf{0.0630} & \textbf{0.9115} & \textbf{0.1751} & 0.1083 & \textbf{0.0909} & \textbf{0.1658}
   & \textbf{0.8660} & 0.9228 & 0.9694\\
   \Xhline{0.8pt}
   \multirow{2}{*}{\makecell{\textit{Fast}\\\textit{Walking}}} & Handcrafted baseline & 0.1473 & 1.2425 & 0.2475 & 0.1508 & 0.1824 & 0.3229 & 0.6839 & 0.8774 & \textbf{0.9702}\\
   \cline{2-11}
   & \textbf{Ours} & \textbf{0.0832} & \textbf{0.9185} & \textbf{0.1870} & \textbf{0.1201} & \textbf{0.1132} & \textbf{0.2093} & \textbf{0.8575} & \textbf{0.9165} & 0.9610\\
   \Xhline{0.8pt}
  \end{tabular}
  \caption{Performance of safety envelope estimation in a real-world dataset with moving pedestrians. The environment consisted of two people walking in both back-and-forth and sideways motions.}
  \label{table:realworld}
\end{table*}

\subsection{Random curtain analysis}
\label{sec:experiments:analysis}

In this section, we use the dynamic programming approach introduced in Section~\ref{sec:random-curtain-analysis:theoretical-analysis} to analyze the detection probability of random curtains. First, we compare our dynamic programming method to an alternate approach to compute detection probabilities: Monte Carlo sampling. This method involves sampling a large number of random curtains and returning the average number of curtains that were able to detect the object. This produces an unbiased estimate of the single-curtain detection probability, with a variance based on the number of samples. However, our dynamic programming approach has multiple advantages over such a sampling-based approach:
\begin{enumerate}
  \item Dynamic programming produces an \textit{analytic} estimate of the detection probability, whereas sampling produces a \textit{stochastic}, noisy estimate of the probability. 
  Analytic estimates 
  %are \textit{probabilistic guarantees}, and 
  are useful for reliably evaluating the safety and robustness of perception systems.
  \item Dynamic programming is significantly more efficient than sampling based approaches. The former only involves one pass through the constraint graph.
  % and the runtime complexity is $O(V + E)$, where $V$ and $E$ are the number of nodes and edges in the graph.
  In contrast, a large number of samples may be required to provide a reasonable estimate of the detection probability.
\end{enumerate}

The two methods are compared in Figure~\ref{fig:analysis:monte-carlo}, which shows the estimated single-curtain detection probabilities of both methods as a function of the runtime of each method \hl{(the runtimes include pre-processing steps such as raycasting, and hence are directly comparable between the two methods)}.
%The X-axis shows the runtime of each method (in seconds), and the Y-axis shows their estimated single-curtain detection probabilities. 
Dynamic programming (shown in red) produces an analytic estimate very efficiently (around 0.8 seconds). For Monte Carlo sampling, we show the probability estimate for a varying number of Monte Carlo samples. Each run shows the mean estimate of the detection probability (blue dots), as well as its corresponding $95\%$-confidence intervals (blue bars). Using more samples produces more accurate estimates with smaller confidence intervals, at the cost of increased runtime. The sampling approach will eventually converge to the point estimate output by dynamic programming in the limit of an infinite number of samples. This experiment shows that dynamic programming produces precise estimates (i.e. there is zero uncertainty in its estimate) while being orders of magnitude faster than Monte Carlo sampling.

Next, we investigate how the size of an object affects the detection probability. We generate objects of varying sizes and run our dynamic programming algorithm to compute their detection probabilities. Figure~\ref{fig:analysis:area-and-curtains} (a) shows a plot of the detection probability of a single curtain as a function of the area of the object (averaged over multiple object orientations). As one would expect, the figure shows that larger objects are detected with higher probability. 
%Note that the orientation of an object can also affect the detection probability; hence, we report 
%These probabilities averaged over multiple orientations. 

Last, we analyze the detection probability as a function of the number of light curtains placed. The motivation for using multiple curtains to detect objects is the following. A single curtain might have a low detection probability $p$, especially for a small object. However, we could place multiple (say $n$) light curtains and report a successful detection if at least one of the  $n$ curtains detects the object. Then, the probability of detection increases exponentially by $1 - (1-p)^n$. We call this the ``multi-curtain" detection probability. Figure~\ref{fig:analysis:area-and-curtains} (b) shows the multi-curtain detection probabilities for objects from the KITTI~\cite{Geiger2012CVPR} dataset, as a function of the time taken to place those curtains (at 60 Hz). For each object class, we construct a ``canonical object" by averaging the dimensions of all labeled instances of that class in the KITTI dataset. We can see that larger object classes are detected with a higher probability, as expected. The figure also shows that the probability increases rapidly with the number of random curtains for all object classes. Four random curtains (which take about 67ms to image) are sufficient to detect objects from all classes with at least $90\%$ probability. 
%This experiment shows that random curtains are able to  detect objects typically encountered in driving  scenarios with very high probability. 
Note that there is a tradeoff between detection probability and runtime of multiple curtains; guaranteeing a high probability requires more time for curtains to be placed.

% We also analyze the detection probability as a function of the number of random light curtains placed. A single light curtain may only be able to detect a small object with a low probability. However, if we place multiple random curtains, the probability that the object will be detected by at least one of the curtains increases exponentially with the number of curtains. The figure shows that the probability increases rapidly with the number of random curtains for all object classes. Four random curtains (which take about 67ms to image for the device running at 60Hz) are sufficient to detect object from all classes with at least $90\%$ probability.

\subsection{Estimating safety envelopes}
\label{sec:experiments:depth-tracking}

\textbf{Environments:} In this section, we evaluate our approach to estimate safety envelopes using light curtains, in two environments. First, we use SYNTHIA~\cite{synthia}, a large, simulated dataset containing photorealistic scenes of urban driving scenarios. It consists of 191 training scenes ($\sim 96K$ frames) and 97 test scenes ($\sim 45K$) frames and provides ground truth depth maps. Second, we perform safety envelope estimation in a real-world environment with moving pedestrians. These scenes consist of two people walking in front of the device in complicated, overlapping trajectories. \hl{We perform evaluations in two settings: a \textit{``Slow Walking"} setting, and a harder \textit{``Fast Walking"} setting where forecasting the motion of the safety envelope is naturally more challenging.} We use an Ouster OS2 128-beam LiDAR (and ground-truth depth maps for the SYNTHIA dataset) to compute ground truth safety envelopes for training and evaluation. We evaluate policies over a horizon of 50 timesteps in both environments.

\textbf{Evaluation metrics: }
Safety envelopes can be thought of as 1D depth maps computed from a full 2D depth map, since the safety envelope is constrained to be a vertically ruled surface that always ``hugs'' the closest obstacle.  Thus, the safety curtain can be computed by selecting the closest depth value along each column of a 2D depth map (ignoring points on the ground or above a maximal height). %(across those pixels whose projected height lies within a pre-specified range to account for ground- and ceiling- subtraction). 
Because of the relationship between the safety envelope and the depth map, we evaluate our method using a variety of standard metrics from the single-frame depth estimation literature~\cite{zhao2020monocular,eigen2014nips}.
%(see Table~(\ref{table:synthia},~\ref{table:realworld}), 
%applied to the special case of 1D depth. 
The metrics are averaged over multiple timesteps to evaluate the policy's performance across time.

\textbf{Baselines:} In Table~\ref{table:synthia}, we compare our method to the hand-crafted policy described in Sec.~\ref{sec:learning-forecasting}.
A `random curtain only' baseline tests the performance of random curtains for safety envelope estimation in the absence of any forecasting policy.
We also compare our method against two other neural network architectures that forecast safety envelopes: a CNN that performs 1D convolutions, and a graph neural network with nodes corresponding to camera rays. Please see Appendix~\ref{sec:appendix:architectures} for more details about their network architectures. See Table~\ref{table:synthia} for a comparison of our method with the baselines in the SYNTHIA environment, and Table~\ref{table:realworld} for the real-world environment. The arrows below each metric in the second row denote whether a higher value ($\uparrow$) or lower value ($\downarrow$) is better. In both environments (simulated and real), our method outperforms the baselines on most metrics, often by a significant margin.

\textbf{Ablations:} We also perform multiple ablation experiments. First, we train and evaluate our without using random curtains \hl{(Tab.~\mbox{\ref{table:synthia}}, ``Ours w/o Random Curtains")}. This reduces the performance by a significant margins, suggesting that it is crucial to combine forecasting with random curtains for increased robustness. \hl{See Appendix~\mbox{\ref{sec:appendix:sim-results-without-random}} for more experiments performed without using random curtains for all the other baselines and ablation conditions.} Second, we perform an ablation in which we train our model without forecasting to the next timestep i.e. the network is only trained to predict the safety envelope of the current timestep (Tab.~\ref{table:synthia}, ``Ours w/o Forecasting"). This leads to a drop in performance, suggesting that it is important to place light curtains at the locations where the safety envelope is expected to move to, not where it currently is. Finally, we modify our method to not take the output of the hand-crafted policy as input (Tab.~\ref{table:synthia}, ``Ours w/o Baseline input"). The drop in performance shows that providing the neural network access to another policy that performs reasonably well helps with training \hl{and improves performance}.

\textbf{Qualitative anaysis}: We perform qualitative analysis of our method in the real-world environment with moving pedestrians in Fig.~\ref{fig:real_world_experiment}, and in the SYNTHIA~\cite{synthia} simulated environment in Figs.~\ref{fig:random_curtain_detects},~\ref{fig:qualitative_synthia}. We compare our method against the hand-crafted baseline, as well as show how placing random curtains can discover objects and improve the estimation of safety envelopes. Please see captions for more details. \hl{Our \mbox{\href{\website}{project website}} contains videos demonstrating the qualitative performance of our method in the real-world pedestrian environment. They show that our method can generalize to multiple obstacles (as many as five pedestrians) and extremely fast and spontaneous motion, even though such examples were not part of the training set.}

% \begin{table*}[h!]
%   \centering
%   \begin{tabular}{?c?c|c|c|c|c|c|c|c|c?} 
%    \Xhline{0.8pt}
%     & Huber loss & \makecell{RMSE\\Linear} & \makecell{RMSE\\Log} & \makecell{RMSE\\Log Scale-Inv.} & \makecell{Absolute\\Relative Diff.} & \makecell{Squared\\Relative Diff.} & \makecell{Thresh\\($1.25$)} & \makecell{Thresh\\($1.25^2$)} & \makecell{Thresh\\($1.25^3$)}\\
%     \hline
%     & $\downarrow$ &  $\downarrow$ & $\downarrow$ & $\downarrow$ & $\downarrow$ & $\downarrow$ & $\uparrow$ & $\uparrow$ & $\uparrow$\\
%     \Xhline{0.8pt}
%     Baseline & 0.1989 & 2.5811 & \textbf{0.2040} & \textbf{0.0904} & 0.2162 & 2.0308 & 0.6321 & 0.7321 & 0.7657\\
%     \hline
%     \makecell{\textbf{Ours}\\(Supervised)} & \textbf{0.1103} & \textbf{2.3508} & 0.2049 & 0.1051 & \textbf{0.1318} & \textbf{0.8354} & \textbf{0.6546} & \textbf{0.7322} & \textbf{0.7709}\\
%    \Xhline{0.8pt}
%   \end{tabular}
%   \caption{Performance of a neural-network based depth-tracking policy trained with supervised learning, using various standard depth estimation metrics~\cite{zhao2020monocular,eigen2014nips}. This is compared to a handcrafted baseline policy.}
%   \label{table:main-results}
%   \vspace{-15pt}
% \end{table*}

\section{Conclusion}
\label{sec:conclusion}

In this work, we develop a method to estimate the safety envelope of a scene, which is a hypothetical vertical surface that separates a robot from all obstacles in the environment. We use light curtains, an actively controllable, resource-efficient sensor  to directly estimate the safety envelope. We describe a method to generate random curtains that respect the physical constraints of the device, in order to quickly discover the safety envelope of an unknown object. Importantly, we develop a dynamic-programming based approach to produce theoretical safety guarantees on the probability of random curtains detecting objects in the scene. We combine this method with a machine-learning based model that forecasts the motion of already-discovered safety envelopes to efficiently track them. This enables our robot perception system to accurately estimate safety envelopes, while our probabilistic guarantees help certify its accuracy and safety towards obstacle detection and avoidance.

\section*{Acknowledgements}
\hl{We thank Adithya Pediredla, N. Dinesh Reddy and Zelin Ye for help with real-world experiments. This material is based upon work supported by the National Science Foundation under Grants No. IIS-1849154, IIS-1900821 and by the United States Air Force and DARPA under Contract No. FA8750-18-C-0092.}

%%%%%%%%%%%%%%%%%%%%%%%%%%%%%%%%%%%%%%%%%%%%%%%%%%%%%%%%%%%%%%%%%%%%%%%%%%%%%%%%

% \section*{Acknowledgments}
% \dave{No acknowledgements in the initial double-blind submission}

%% Use plainnat to work nicely with natbib. 
\newpage
\bibliographystyle{plainnat}
\bibliography{references}

\clearpage
\appendix

\subsection{Transition distributions for sampling random curtains}
\label{sec:appendix:sampling-curtains}

We first describe, in Algorithm~\ref{alg:sampling-curtain} the procedure to sample a random curtain from the extended constraint graph $\Ge$ by successively generating it using a transition probability function $P(\X_{t+1} \mid \X_{t-1}, \X_t)$. The only constraint on $P(\X_{t+1} \mid \X_{t-1}, \X_t)$ is that it must equal to $0$ if $\X_{t-1}, \X_t, \X_{t+1}$ do not satisfy both the velocity and acceleration constraints of Equations~(\ref{eqn:vel-constraint},~\ref{eqn:acc-constraint}).

\begin{algorithm}
  \caption{Sampling a random curtain from $\Ge$}
  \DontPrintSemicolon
  \tcc{Inputs}
  $\Ge \gets$ extended constraint graph\;
  $P(\X_{t+1} \mid \X_{t-1}, \X_t) \gets$ transition prob. distribution\;
  $P((\X_1, \X_2)) \gets$ initial probability distribution\;\;
  
  \tcc{Progressively generate curtain}
  Curtain $\gets \{\}$\;\;

  \tcc{Initialization}
  Sample $(\X_1, \X_2) \sim P((\X_1, \X_2))$\;
  Curtain $\gets \{\X_1, \X_2\}$\;\;
  
  \tcc{Iteration}
  \For{$t = 2$ \text{to} $T-1$}{
      $\X_{t+1} \sim P(\X_{t+1} \mid \X_{t-1}, \X_t)$\;
      Curtain $\gets \text{Curtain} \cup \{\X_{t+1}\}$
   }\;  
   \KwRet Curtain
  \label{alg:sampling-curtain}
\end{algorithm}

We initialize by sampling a location $\Ne_2 = (\X_1, \X_2)$ according to an initial sampling distribution. At the $(t-1)$-th iteration, we will have sampled the $t-1$ nodes $(\Ne_2, \dots, \Ne_t)$ corresponding to the first $t$ control points $(\X_1, \dots, \X_t)$ of the random curtain. At the $t$-th iteration, we sample $\X_{t+1}$ according to the transition probability distribution $\X_{t+1} \sim P(\X_{t+1} \mid \X_{t-1}, \X_t)$ and add $\X_{t+1}$ to the current set of control points. After all iterations are over, this algorithm generates a complete random curtain.

The above procedure to sample random curtain provides the flexibility to design any initial and transition probability distribution functions. Then, what are good candidate distributions? We use random curtains to detect the presence of objects in the scene whose location is unknown. Hence, the objective is to find the light curtian sampling distribution that maximizes the probability of detection of an object that might be placed at any arbitrary location in a scene. This objective will be achieved by random curtains that \textit{cover a large area}. We now discuss a few sampling methods and qualitatively evaluate them in terms of the area covered by random curtains generated from them.

\begin{figure*}[t]
  \centering
  \subfloat[Uniform sampling of neighbors]{
      \includegraphics[width=0.33\textwidth,trim=0 0 0 0]{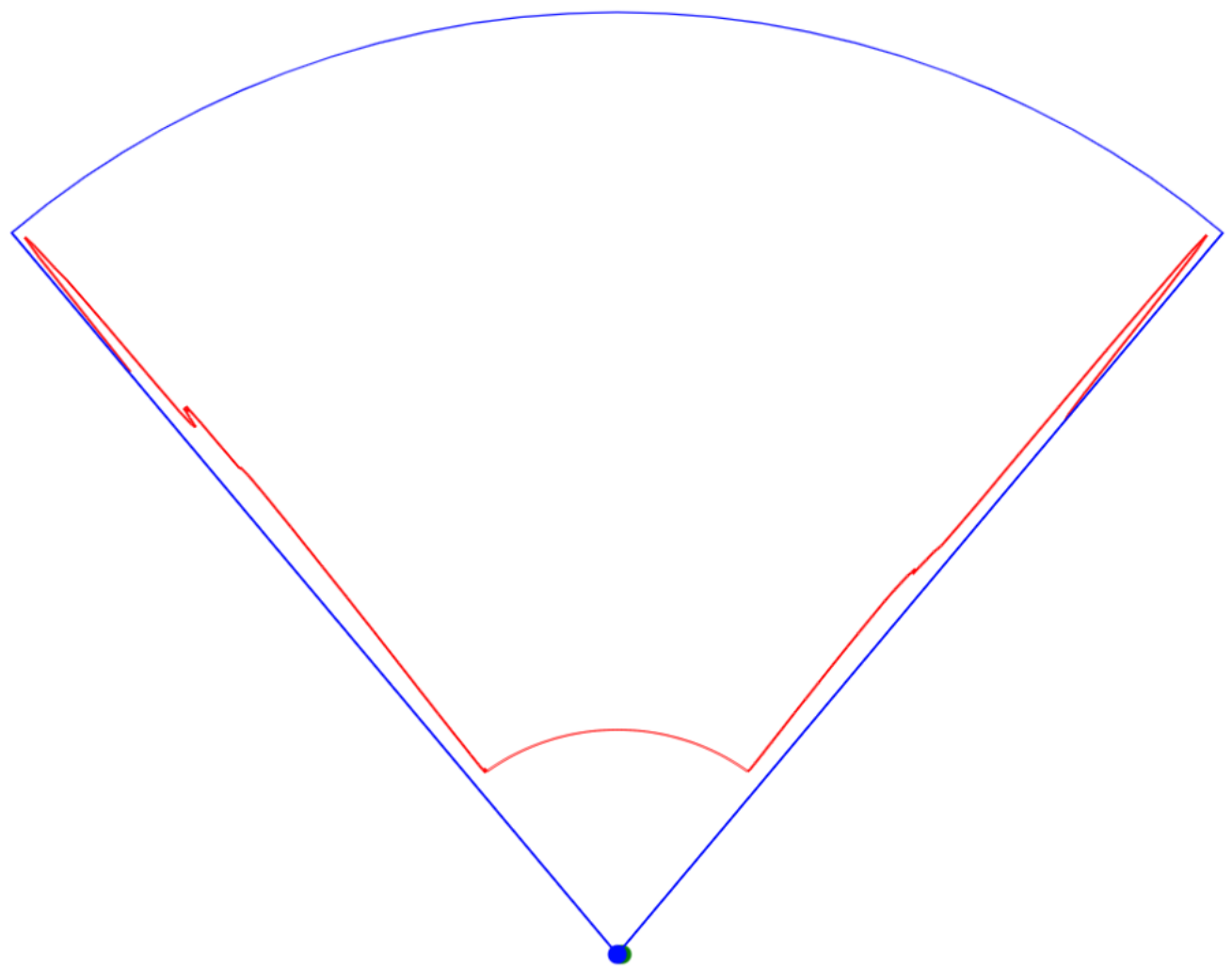}
  }
  \subfloat[Uniform linear sampling of setpoints]{
      \includegraphics[width=0.33\textwidth,trim=0 0 0 0]{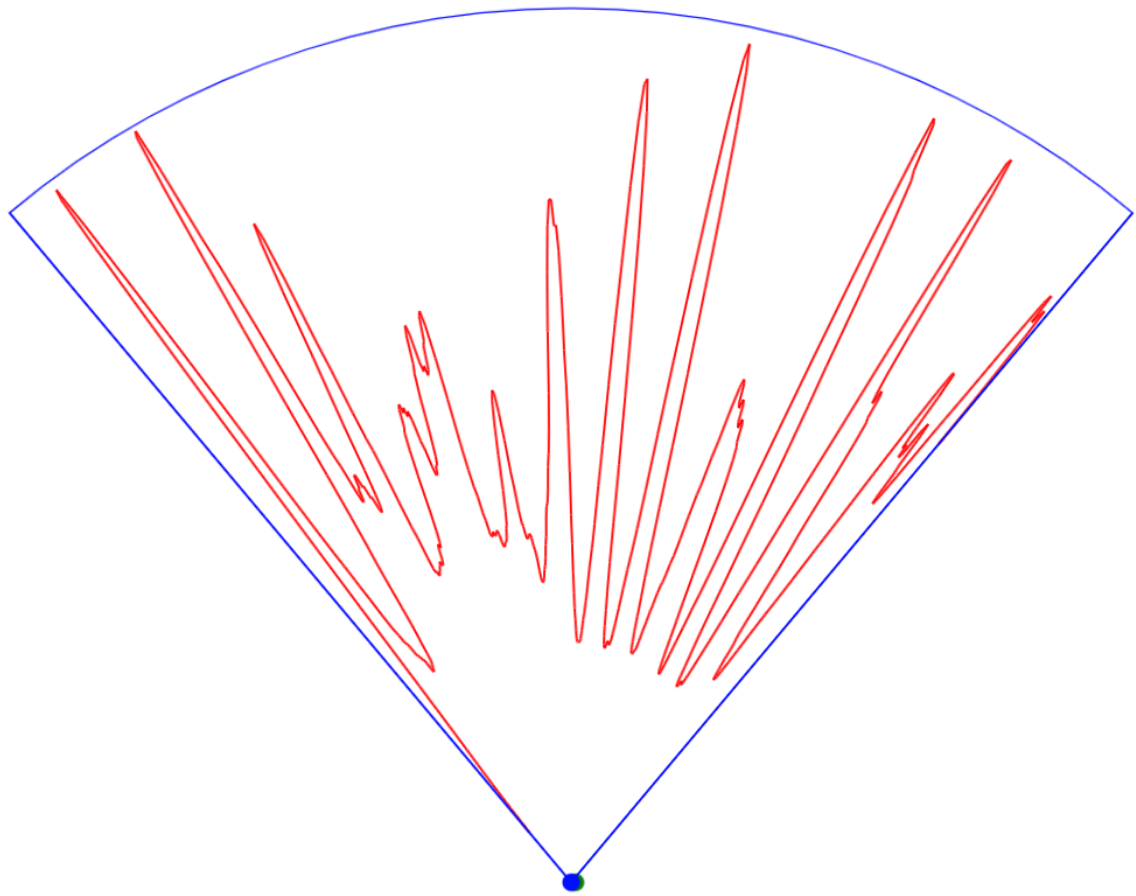}
  }
  \subfloat[Uniform area sampling of setpoints]{
      \includegraphics[width=0.33\textwidth,trim=0 0 0 0]{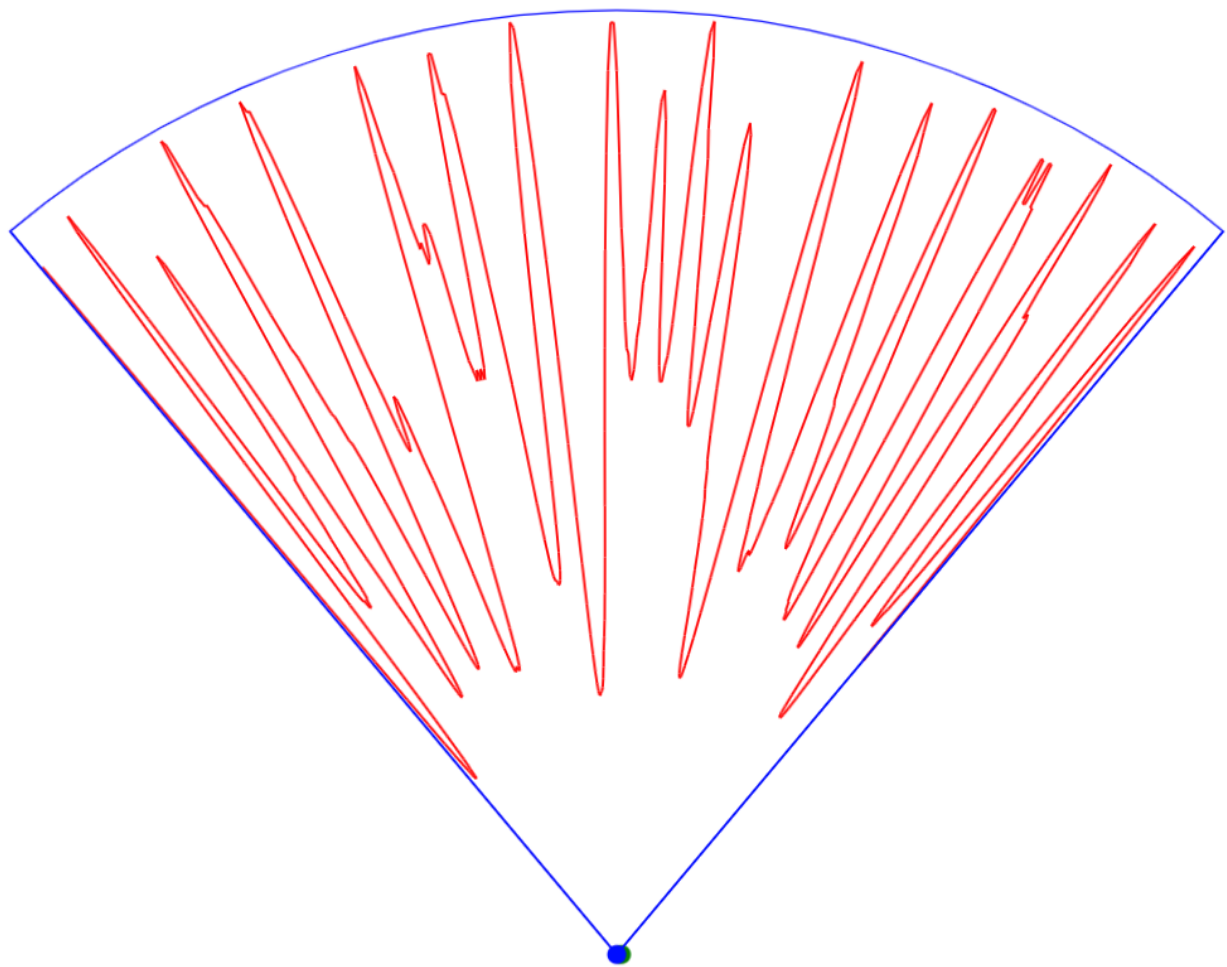}
  }
  \caption{Qualitative comparision of the coverage of random light curtains under different transition probability distributions. Sampled random curtains are shown in red. (a) \textit{Uniform neighbor sampling}: for a given node, its neighbors on the next camera ray are sampled uniformly at random. This can produce random curtains that are at a constant distance away from the device. (b) \textit{Uniform linear setpoint sampling}: for every camera ray, a setpoint distance $r \in [0, r_\mathrm{max}]$ is sampled uniformly at random. Then the neighbor closest to the setpoint is chosen. This has significantly higher coverage, but is biased towards sampling locations close to the device. (c) \textit{Uniform area setpoint sampling}: for every camera ray, a setpoint distance $r \in [0, r_\mathrm{max}]$ is sampled with a probability proportional to $r$. This assigns a higher probability to a larger $r$, and corresponds to uniform \textit{area} sampling. Then the neighbor closest to the setpoint is chosen. This method qualitatively exhibits the best coverage.}
  \label{fig:rc-sampling}
\end{figure*}

\textbf{1. Uniform neighbor sampling}: Perhaps the simplest transition probability distribution $P(\X_{t+1} \mid \X_{t-1}, \X_t)$ for a given extended node $(\X_{t-1}, \X_t)$ is to select a neighboring control point $\X_{t+1}$ (that is connected by a valid edge originating from the node) with uniform probability. However, since the distribution does not take into account the physical locations of the current control point, it does not explicitly try to maximize coverage. To illustrate this, consider a random curtain that starts close to light curtain device. If it were to maximize coverage, the galvanometer would need to rotate so that the light curtain is placed farther from the device on subsequent camera rays. However, since its neighboring nodes are selected at random, the sampled locations on the next ray are equally likely to be nearer to the device than farther away from it. This can produce random curtains as shown in Fig.~\ref{fig:rc-sampling} (a).\\

\textbf{2. Uniform linear setpoint sampling}: a more principled way to sample neighbors is inspired by rapidly-exploring random trees (RRTs), which are designed to quickly explore and cover a given space. During tree expansion, an RRT first randomly samples a \textit{setpoint} location, and selects the vertex that is closest to that location. We adopt a similar procedure. For any current node $(\X_{t-1}, \X_t)$, we first sample a setpoint distance $r \in [0, r_\mathrm{max}]$ uniformly at random on a line along the camera ray. The probability density of $r$ is a constant and equal to $P(r) = 1 / r_\mathrm{max}$. Then, we select a valid neighbor $\X_{t+1}$ that is closest to this setpoint location among all valid neighbors.
% This procedure is inspired by rapidly-expanding random-trees (RRT), which are designed to quickly explore and cover a given space. During tree expansino, an RRT first randomly samples a setpoint location, and choses the vertex that is closest to that location. RRTs a
Let us again consider the situation described in the previous approach, where the current node is located close to the light curtain device. When a setpoint is sampled uniformly along the next camera ray, there is high probability that it will correspond to a location that is farther away from the current node. Hence, the neighboring location $\X_{t+1}$ that is chosen on the next ray will likely lie away from the device as well. A random curtain sample generated from this distribution is shown in Fig.~\ref{fig:rc-sampling} (b). It tends to alternate between traversing near regions and far regions of the space in front of the curtain, covering a larger area than the previous sampling approach.\\

\textbf{3. Uniform area setpoint sampling}: We use the setpoint sampling approach described above, but revisit how the setpoint itself is sampled. Previously, the setpoint $r \in [0, r_\mathrm{max}]$ was sampled uniformly at random on the line along the ray, with $P(r) = 1/r_\mathrm{max}$. Now, we propose an alternate sampling distribution and provide some theoretical justification. Since we want to uniformly cover the area in front of the light curtain device, consider the following experiment. Let us sample a point $(x, y)$ uniformly from the area within a circle of radius $r_\mathrm{max}$ (or equivalently, within any sector of that circle). Then the cumulative distribution function of $r = \sqrt{x^2 + y^2}$ is $P(r < r') = \pi r'^2 /\pi r_\mathrm{max}^2$ (area of the smaller circle divided by the area of entire circle), which implies that the probability density function of $r$ is equal to $P(r) = 2r / r_\mathrm{max}^2$. This suggests that we must assign a higher probability to a larger $r$ (proportional to $r$), since a larger area exists away from the center of the circle than near the center. Hence, we sample the setpoint $r$ from $P(r) = 2r / r_\mathrm{max}^2$, by first sampling $s \sim \text{Uniform}(0, r_\mathrm{max}^2)$ and then setting $r = \sqrt{s}$. Finally, we select the valid neighbor $\X_{t+1}$ on the next ray that is closest to this setpoint. Since this method is motivated by sampling areas rather than sampling along a line, we call this approach ``area setpoint sampling''. An example curtain sampled using this approach is visualized in Fig.~\ref{fig:rc-sampling}~(c), which generally exhibits the best coverage among all methods. We use this method to sample random light curtains for all experiments in this paper.

% In practice, we found that the best results \dave{best according to what metric? what is the objective here?} were obtained when we use a $\ptrans(\Ne_{t+1} \mid \Ne_t)$ that  corresponds to \dave{use a Ptrans that corresponds to $\rightarrow$ define Ptrans as (write a mathematical definition for Ptrans)} first randomly sampling a target $\X_\mathrm{target}^2 \sim \text{Uniform}(0, \X_\mathrm{max}^2)$ \dave{why is everything squared?} and picking the node $\Ne_{t+1}-{(k)}$ closest to $\X_\mathrm{target}$, i.e. $\arg \min_k |\Xcurr(\Ne_{t+1}-{(k)}) - \X_\mathrm{target}|$ \dave{You should clarify that you set $X_prev$ to be the value chosen at the previous iteration of the algorithm, so really $P_trans$ is just a distribution over $X_t$ not $N_t$ (defining it as a distribution over $X_t$ would probably simplify things)}.
\subsection{Dynamic programming for computing detection probability}
\label{sec:appendix:dp}

\begin{algorithm}
  \caption{Dynamic programming to compute detection probabilities $\Ge$}
  \DontPrintSemicolon
  \tcc{Inputs}
  $\Ge \gets$ extended constraint graph\;
  $P(\X_{t+1} \mid \X_{t-1}, \X_t) \gets$ transition prob. distribution\;
  $P((\X_1, \X_2)) \gets$ initial probability distribution\;
  $\{O_1, \dots, O_T\} \gets $ ground truth object locations\;
  $\tau \gets $ detection intensity threshold\;\;
  
  $\set_t \gets$ the set of nodes on the $t$-th camera ray in the constraint graph.\;\;

  \tcc{Detection at each location}
  \ForAll{$\X_t \in \set_t,\ \ 1 \leq t \leq T$}{
    $\I_t(\X_t \mid O_t) \gets $ intensity using rendering\;   
    $\dd_t(\X_t \mid O_t) \gets [\I_t(\X_t \mid O_t) > \tau]$\;
  }\;
  
  \tcc{Initializing last ray}
  \For{$(\X_{T-1}, \X_T) \in \set_T$}{
    $\pdet(\X_{T-1}, \X_T) \gets \dd_t(\X_T \mid O_T)$
  }\;

  \tcc{Dynamic programming loop}
  \For{$t = T-1$ \text{to} $2$}{
    \For{$(\X_{T-1}, \X_T) \in \set_t$}{
      \If{$\dd_t(\X_t, O_t) = 1$} {
        $\pdet(\X_{T-1}, \X_T) \gets 1$
      }
      \Else{
        $\pdet(\X_{T-1}, \X_T) \gets \sum_{\X_{T+1}} \pdet(\X_T, \X_{T+1}) \cdot P(\X_{t+1} \mid \X_{t-1}, \X_t)$
      }
    }
  }\;

  \tcc{Initial ray}
  $P(\dd) \gets 0$\;
  \For{$(\X_1, \X_2)$}{
    $P(\dd) \gets P(\dd) + P(\X_1, \X_2) \cdot$\;
    $\phantom{P(\dd) \gets}\pdet(\X_1, \X_2)$
  }\;
   
   \KwRet $P(\dd)$
  \label{alg:sampling-curtain}
\end{algorithm}

To compute the quantity $P(\dd)$, we first decompose the overall problem into smaller subproblems by defining the \textit{sub-curtain detection probability} $\pdet(\X_{t-1}, \X_t) = P(\bigvee_{t'=t}^T \dd_t \mid \X_{t-1}, \X_t)$. This is the probability that a random curtain \textit{starting} on the extended node $(\X_{t-1}, \X_t)$ and ending on the last camera ray, detects the object $O$ between rays $\R_t$ and $\R_T$. Note that the overall detection probability can be written in terms of the sub-curtain detection probabilities of the second ray as $P(\dd) = \sum_{\set_2} P(\X_1, \X_2) \cdot \pdet(\X_1, \X_2)$. Then we iterate over camera rays from $\R_T$ to $\R_2$. The node detection probabilities on the last ray will simply be either $1$ or $0$, based on whether the object is detected at the node or not. After having computed node detection probabilities for all rays between $t+1$ and $T$, the probabilities for nodes at ray $t$ can be computed using a recursive formula. Finally, after obtaining the probabilities for nodes on the initial rays, the overall detection probabilities can be computed as described previously.
\subsection{Computational complexity of the extended constraint graph}
\label{sec:appendix:runtime}

\hl{
In this section, we discuss the computational complexity associated with the extended constraint graph. Let $K$ be the number of discretized control points per camera ray in the constraint graph, and let $T$ be the number of camera rays.

\textbf{Constraint graph size:} in the original constraint graph $\Go$ of \mbox{\citet{ancha20eccv}}, since a node $\No_t = \X_t$ contains only one control point, there can be $O(K)$ nodes per camera ray and $O(K^2)$ edges between consecutive camera rays. This means that there are $O(TK)$ nodes and $O(T K^2)$ edges in the graph.

However, in the extended constraint graph $\Ge$, each node $\Ne_t = (\X_{t-1}, \X_t)$ contains a pair of control points. Hence, there can be up to $O(K^2)$ nodes per camera ray and $O(K^4)$ edges between consecutive camera rays! This implies that the total nodes and edges in the graph can be up to $O(T K^2)$ and $O(T K^4)$ respectively.

\textbf{Dynamic programming:} dynamic programming involves visiting each node and each edge in the graph once. Therefore, the worst-case computation time of dynamic programming in the extended constraint graph, namely $O(TK^4)$, might seem prohibitively large at first. However, the additional acceleration constraints in $\Ge$ can significantly limit the increase in the number of nodes and edges. Additionally, we perform graph pruning as a post-processing step, to remove all nodes in the graph that do not have any edges. Since the topology of the constraint graph is fixed, the graph creation and pruning steps can be done offline and only once. These optimizations enable our dynamic programming procedure to be very efficient, as shown in Sec.~\mbox{\ref{sec:experiments:analysis}}. That being said, any slow down in dynamic programming is generally acceptable because it is only used for offline probabilistic analysis.

\textbf{Random curtain generation:} random curtains are placed by our online method. Fortunately, generating random curtains from the constraint graph is very fast. It involves a single forward pass (random walk) through the graph, visiting exactly one node per ray. It also involves parsing each visited node's transition probability distribution vector, whose length is equal to the number of edges of that node. Since both $\Go$ and $\Ge$ can have at most $K$ edges per node, the runtime of generating a random curtain is $O(TK)$ (for both $\Go$ and $\Ge$). In practice, a large number of random curtains can be precomputed offline.
}

% Note that incorporating acceleration constraints in the constraint graph by extending its nodes may lead to an explosion in size of the graph. Let us assume that we use $N$ discretized locations per camera ray. The constraint graph of \citet{ancha20eccv} can contain up to $N^2$ edges between two consecutive rays. When using our extended nodes, there can be up to $N^2$ nodes per ray and up to $N^4$ edges between rays! However, the additional acceleration constraints can limit the increase in the number of edges. Additionally, we perform graph pruning to limit the size of the extended graph. It is possible for certain extended nodes to be associated with such large velocities that they will always drive the light curtain outside the working range of the device (approximately 20m). We remove such nodes (and corresponding edges) from the graph if there exists no path starting from that node, and ending on the last ray, that entirely lies within the working range. \dave{How can this be?}\sid{Added a clarification above. Since all discretized locations in the graph are created to lie within the maximum working range of the light curtain (20m), saying that a ``path exists to the last ray of the graph" is the same as saying that a ``path exists that lies entirely within the working range"}. Since the topology of the constraint graph is fixed, the graph creation and graph pruning steps can be done offline and only once. Hence, they do not impact the online runtime of our algorithms.

\subsection{Network architectures and training details}
\label{sec:appendix:architectures}

In this section, we describe in detail the network architectures used by our main method, as well as various baseline models.

\subsubsection{\textbf{2D-CNN}}

\begin{figure*}[t]
    \centering
    \includegraphics[scale=.2,trim=0 0 0 0]{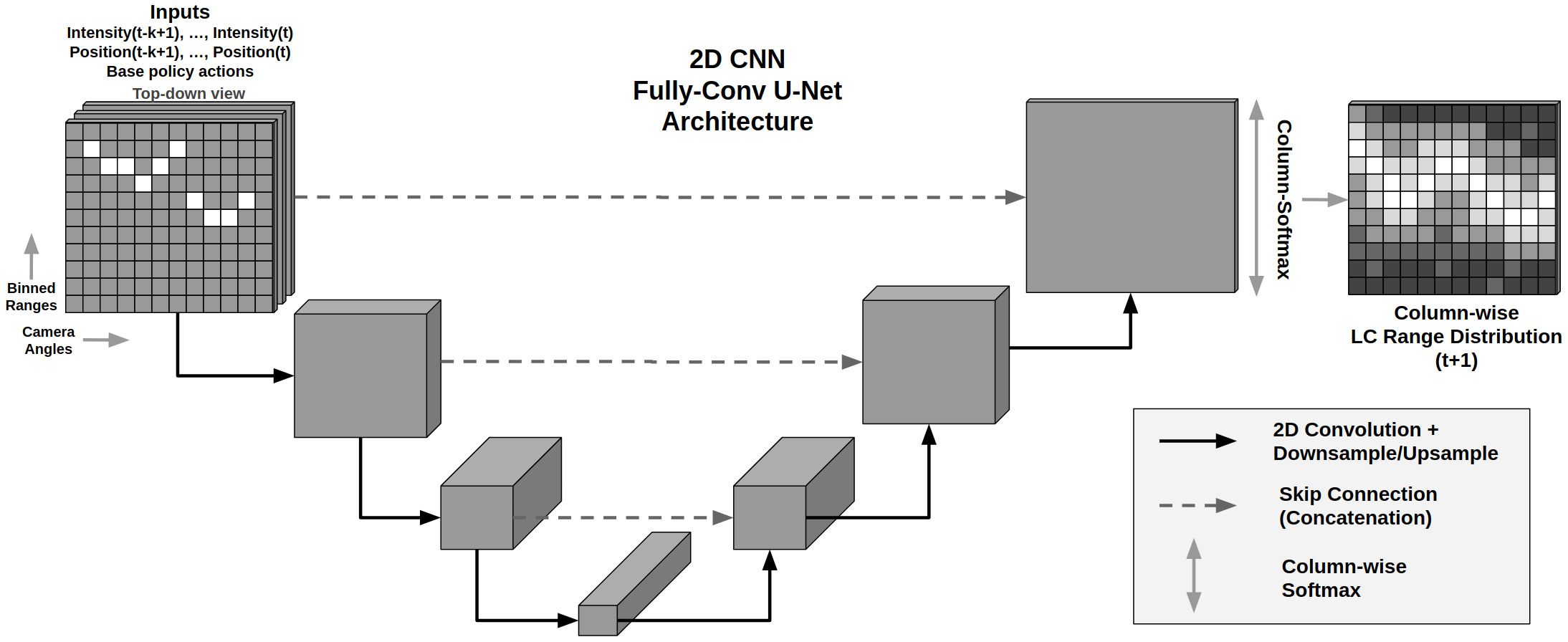}
    \caption{The network architecture of the  2D CNN model used for safety envelope forecasting. It takes as input the previous $k$ light curtain outputs, and converts them into top-down polar occupancy maps. Each column of the image is assigned to a camera ray, and each row is treated as a binned location. It also takes the prediction of the hand-crafted baseline as additional input. The input is transformed through a series of 2D convolution layers, arranged in a manner similar to the U-Net~\cite{ronneberger2015u} architecture. This involves skip connections between downsampled and upsampled layers with the same spatial size. The output of the U-Net is a 2D image. This is a fully-convolutional architecture, and the spatial dimensions of the input and output are equal. Column-wise soft-max is then applied to the output to transform it to a probability distribution per column. A value $\X_t$ is sampled per column to produce the profile of the forecasted safety envelope.}
    \label{fig:cnn2d}
  \end{figure*}

The 2D-CNN architecture we use to forecast safety envelopes is shown in Fig.~\ref{fig:cnn2d}. It takes as input the previous $k$ light curtain outputs. These consists of the intensities of the light curtain per camera ray $\I_{1:T}$, as well as the control points of the curtain that was placed i.e. $\X_{1:T}$. Each light curtain output $(\X_{1:T}, \I_{1:T})$ is converted into a \textit{polar occupancy map}. A polar occupancy map is a $T \times L$ image, where the $t$-th column of the image corresponds to the camera ray $\R_t$. Each ray is binned into $L$ uniformly spaced locations; although $L$ could be set to the number of control points per camera ray in the light curtain constraint graph, it is not required. Each column of the occupancy map has at-most one non-zero cell value. Given $\X_t, \I_t$, the cell on the $t$-th column that lies closest to $\X_t$ is assigned the value $\I_t$. We generate $k$ such top-down polar occupancy maps encoding intensities. We generate $k$ more such polar occupancy maps, but just assigning binary values to encode the control points of the light curtain. Finally, another polar occupancy map is generated using the forecast of the safety envelope from the handcrafted baseline policy. The $2k+1$ maps are fed as input to the 2D-CNN. We use $k=5$ in all our experiments. The input is transformed through a sequence of 2D convolutions; the convolutional layers are arranged in a manner similar to the the U-Net~\cite{ronneberger2015u} architecture. This involves skip connections between downsampled and upsampled layers with the same spatial size. The output of the U-Net is a 2D image. The U-Net is a fully convolutional architecture, and the spatial size of the output is equal to the spatial size of the input. Column-wise soft-max is then applied to transform the output into $T$ categorical probability distributions, one per column. We sample a cell from the $t$-th distribution, and the location of that cell in the top-down view is interpreted as the $t$-th control point. This produces a forecasted safety envelope.   

\subsubsection{\textbf{1D-CNN}}

We use a 1D-CNN as a baseline network architecture. The 1D-CNN takes as input the previous $k$ light curtain placements $\X_{1:T}$, and treats it as a 3-channel 1-D image (the three channels being the x-coordinate, the z-coordinate, and the range $\sqrt{x^2 + z^2}$). It also takes the previous $k$ intensity outputs $\I_{1:T}$, and treats them as 1-dimensional vectors. It also takes as input the forecasted safety envelope from the hand-crafted baseline. The overall input to the 1D-CNN is a $4k+1$ channel 1D-image. It applies a series of 1D fully-convolutional operations, with ReLU activations. The output is a 1-D vector of length $T$. These are treated as ranges on each light curtain camera ray, and are converted to the control points $\X_{1:T}$ of the forecasted safety envelope.

\subsubsection{\textbf{1D-GNN}}

We use a graph neural network as a baseline to perform safety envelope forecasting. The GNN takes as input the output of the previous two light curtain placements. The GNN contains $2T$ nodes, $T$ nodes corresponding to each curtain. The graph contains two types of edges: vertical edges between corresponding nodes of the two curtains ($T$ in number), and horizontal edges between nodes corresponding to adjacent rays of the same curtain ($2T-1$ in number). Each node gets exactly one feature: the intensity value of its corresponding curtain and camera ray. Each horizontal and vertical edge gets $3$ input features: the differences in the $x, z, \sqrt{x^2 + z^2}$ coordinates of the control points of the rays corresponding to the nodes the edge is connected to. Then, a series of graph convolutions are applied. The features after the final graph convolution, on the nodes corresponding to the most recent light curtain placement are treated as range values on each camera ray $\R_t$. The $t$-th range value is converted to a control point $\X_t$ for camera ray $\R_t$, and the GNN generates a forecast $\X_{1:T}$ of the safety envelope.

\hl{We find that providing the output of the hand-crafted baseline policy as input to the neural networks improves performance (compare the last two rows of Table~\mbox{\ref{table:synthia}}). We attribute this improvement to two reasons:}
\begin{enumerate}
    \item \hl{It helps \textit{avoid local minima during training}: when training the neural networks without the handcrafted input, we observe that the networks quickly settle into local minima where the loss is unable to decrease significantly. This suggests that the input helps with training.}
    \item \hl{It \textit{provides useful information to the network}: To determine if it is also useful after training is complete, we replace the handcrafted input with a constant value and find that this significantly deteriorates performances. This indicates that the 2D CNN continues to rely on the handcrafted inputs at test time.}
\end{enumerate}

\subsection{Parallelized pipelining and runtime analysis}
\label{sec:appendix:pipeline}

\begin{figure*}
    \centering
    \includegraphics[width=0.9\textwidth]{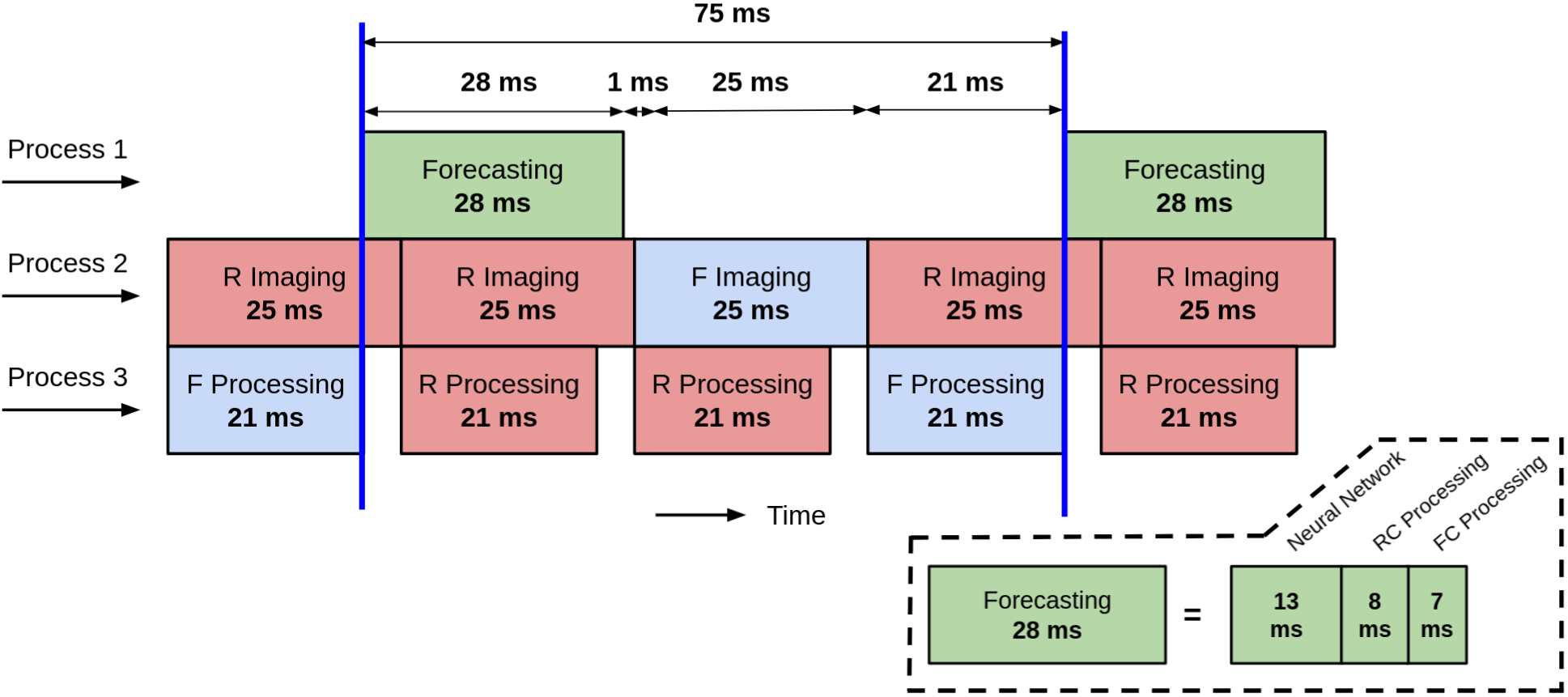}
\caption{\hl{Pipeline showing the runtime of the efficient, parallelized implementation of our method. The pipeline contains three processes running in parallel: (1) the method that forecasts the safety envelope, (2) imaging of the light curtains performed by the physical light curtain device, and (3) low-level processing of images. Here, ``R" or ``RC" stands for ``random curtain" and ``F" or ``FC" stands for ``forecasting curtain". \textit{Bottom right:} the forecasting method further consists of running the feed-forward pass of the 2D CNN and high-level processing of the random and forecasting curtains. The overall latency of our pipeline is 75ms (13.33 Hz). We are able to place two light curtains in each cycle of the pipeline.}}
\label{fig:pipeline}
\end{figure*}

\hl{
In this section, we describe the runtime of our approach. Our overall pipeline has three components: (1) \textit{forecasting} the safety envelope, (2) \textit{imaging} the forecasted and random light curtains, and (3) \textit{processing} the light curtain images. Since these processes can be run independently, we implement them as parallel threads that run simultaneously. This is shown in Figure~\mbox{\ref{fig:pipeline}}.

The imaging and processing threads run continuously at all times. If a forecasted curtain is available to be imaged, it is given priority and is scheduled for the next round of imaging. But if there are no forecasted curtain waiting to be imaged, random curtains are placed and processed. This scheduling leads to an overall latency of 75ms (13.33 Hz). Due to the parallelized implementation, we are able to place two random curtains during each cycle of our pipeline.

Figure~\mbox{\ref{fig:pipeline}} (\textit{right}) shows a breakdown of the timing of the forecasting method method. It consists of the feed-forward pass of the 2D CNN, as well as other high-level processing tasks.
}

\subsection{Results for the simulated environment without using random curtains}
\label{sec:appendix:sim-results-without-random}

\hl{
In this section, we include additional results corresponding to the ``Ours w/o Random curtains'' row of Table~\mbox{\ref{table:synthia}}.
% The row contains results of our method without using random curtains.
Table~\mbox{\ref{table:synthia-full}} contains results of other policies (handcrafted baseline, 1D-CNN baseline, 1D-GNN baseline) and ablation conditions (Ours w/o Forecasting, Ours w/o Baseline input) when random curtains are not used. The top half of Table~\mbox{\ref{table:synthia-full}} contains results without using random curtains. The bottom half contains results for the same policies using random curtains (this is essentially a copy of Table~\mbox{\ref{table:synthia}}, to aid with comparisions). We find that the conclusions of Table~\mbox{\ref{table:synthia}} still hold when random curtains are not used: our method still outperforms the baselines and removing any component of our method (not forecasting to the next timestep or removing the output of the hand-crafted policy as input) reduces performance.
}

\begin{table*}
    \centering
    \begin{tabular}{?c?c?c|c|c|c|c|c|c|c|c?} 
     \Xhline{0.8pt}
       & \multirow{2}{*}{\makecell{Random\\curtain}} & Huber loss & \makecell{RMSE\\Linear} & \makecell{RMSE\\Log} & \makecell{RMSE\\Log Scale-Inv.} & \makecell{Absolute\\Relative Diff.} & \makecell{Squared\\Relative Diff.} & \makecell{Thresh\\($1.25$)} & \makecell{Thresh\\($1.25^2$)} & \makecell{Thresh\\($1.25^3$)}\\
       \cline{3-11}
      & & $\downarrow$ &  $\downarrow$ & $\downarrow$ & $\downarrow$ & $\downarrow$ & $\downarrow$ & $\uparrow$ & $\uparrow$ & $\uparrow$\\
      \Xhline{0.8pt}
       Handcrafted baseline & \xmark & 0.1989 & 2.5811 & 0.2040 & 0.0904 & 0.2162 & 2.0308 & 0.6321 & 0.7321 & 0.7657\\
       \hline
      1D-CNN & \xmark & 0.1522 &  2.3856 & 0.2176 & 0.1076 & 0.1750 & 0.9482 & 0.5842 & 0.7197 & \textbf{0.7868}\\
      \hline
      1D-GNN & \xmark & 0.1584 & 2.2114 & 0.1835 & \textbf{0.0839} & 0.1772 & 1.1999 & 0.6546 & 0.7381 & 0.7710\\
      \hline
      Ours w/o Forecasting & \xmark & 0.1691 & 2.6047 & 0.2288 & 0.1158 & 0.1927 & 1.2555 & 0.6109 & 0.7114 & 0.7654\\
      \hline
      Ours w/o Baseline input & \xmark & 0.1556 & 2.5987 & 0.2273 & 0.1135 & 0.1797 & 1.1063 & 0.6021 & 0.7094 & 0.7683\\
      \hline
      % Ours w/o 5 frames & \xmark & & & & & & & & &\\
      % \hline
      \textbf{Ours} & \xmark & \textbf{0.1220} & \textbf{2.0332} & \textbf{0.1724} & 0.0888 &  \textbf{0.1411} & \textbf{0.9070} & \textbf{0.6752} & \textbf{0.7450} & 0.7852\\
     \Xhline{0.8pt}
     Handcrafted baseline & \cmark & 0.1145 & 1.9279 & 0.1522 & 0.0721 & 0.1345 & 1.0731 & 0.6847 & 0.7765 & 0.8022\\
      \hline
      Random curtain only & \cmark & 0.1484 & 2.2708 & 0.1953 & 0.0852 & 0.1698 & 1.2280 & 0.6066 & 0.7392 & 0.7860\\
       \hline
      1D-CNN & \cmark & 0.0896 & 1.7124 & 0.1372 & 0.0731 & 0.1101 & 0.7076 & 0.7159 & 0.7900 & 0.8138\\
      \hline
      1D-GNN & \cmark & 0.1074 & 1.6763 & 0.1377 & 0.0669 & 0.1256 & 0.8916 & 0.7081 & 0.7827 & 0.8037\\
      \hline
      Ours w/o Forecasting & \cmark & 0.0960 & 1.7495 & 0.1428 & 0.0741 & 0.1163 & 0.6815 & 0.7010 & 0.7742 & 0.8024\\
      \hline
      Ours w/o Baseline input & \cmark & 0.0949 & 1.8569 & 0.1600 & 0.0910 & 0.1148 & 0.7315 & 0.7082 & 0.7740 & 0.7967\\
      \hline
      % Ours w/o 5 frames & \cmark & & & & & & & & &\\
      % \hline
      \textbf{Ours} & \cmark & \textbf{0.0567} & \textbf{1.4574} & \textbf{0.1146} & \textbf{0.0655} & \textbf{0.0760} & \textbf{0.3662} & \textbf{0.7419} & \textbf{0.8035} & \textbf{0.8211}\\
     \Xhline{0.8pt}
    \end{tabular}
    \caption{Performance of safety envelope estimation on the SYNTHIA~\cite{synthia} urban driving dataset under various metrics, with and without using random curtains. Policies in the top half of the table were trained and evaluated without random curtains, while policies in the bottom half were trained and evaluated with random curtain placement.}
    \label{table:synthia-full}
  \end{table*}
\subsection{Hardware specification of light curtains}
\label{sec:appendix:lc-specs}

In this section, we provide some details about the hardware specification of light curtains, as well as comparing it with the specifications of a Velodyne HDL-64E LiDAR.

The distance between the camera and the laser (the baseline of the device) is $20$ cm. The maximum angular velocity of the galvanometer is
$2.5 \times 10^4$
rad/sec and the maximum angular acceleration of the galvanometer is $1.5 \times 10^7$ rad/sec\textsuperscript{2}. The operating range of the light curtain device is up to $20$ meters (daytime outdoors) and $50$ or more meters (indoor or night time).

The following table compares the light curtain device with a Velodyne HDL-64E:

\begin{table}
    \centering
    \begin{tabular}{?c?c|c?} 
     \Xhline{0.8pt}
      & Light curtain & Velodyne HDL-64E LiDAR\\
      \Xhline{0.8pt}
      Horizontal resolution & $0.08\degree$ & $0.08\degree$ -– $0.35\degree$\\
      \hline
      Vertical resolution & $0.07\degree $ & $0.4\degree$\\
      \hline
      Rotation speed & $60$ Hz & $5$ Hz -- $30$ Hz\\
      \hline
      Cost & Less than \$$1000$ &  Approx. \$$80,000$\\
      \Xhline{0.8pt}
    \end{tabular}
    \caption{Performance of safety envelope estimation in a real-world dataset with moving pedestrians. The environment consisted of two people walking in both back-and-forth and sideways motions.}
    \label{table:lc-specs}
  \end{table}

  The LiDAR is limited to fixed scan patterns. Light curtains are designed to be programmable as long as the curtain profiles satisfy the velocity and acceleration limits. Note that the resolution of the light curtain is the same as the 2D camera used which can be significantly higher than any LIDAR. Our current prototype uses a camera with a resolution of $640 \times 512$.

\subsection{Results for the real-world environment under high latency}
\label{sec:appendix:previous-results}

\hl{
The results for the real-world enviroment with walking pedestrians (see Table~\mbox{\ref{table:realworld}} of Section~\mbox{\ref{sec:experiments}}) were generated using the parallelized and efficient pipeline described in Appendix~\mbox{\ref{sec:appendix:pipeline}}. We now present some older results for the same environment that did not use the efficient implementation. Random curtains were not imaged and processed in parallel with the forecasting method. Instead, these operations were performed sequentially: we alternated between the forecasting step and placing a single random curtain. This increases the latency of the pipeline. A comparison between our method and the handcrafted baseline when both use this slower implementation is shown in Table~\mbox{\ref{table:high-latency}}. Our method is able to outperform the handcrafted baseline under various implementations with varying latencies.
}

\begin{table*}[h!]
\centering
\begin{tabular}{?c?c|c|c|c|c|c|c|c|c?} 
    \Xhline{0.8pt}
    & Huber loss & \makecell{RMSE\\Linear} & \makecell{RMSE\\Log} & \makecell{RMSE\\Log Scale-Inv.} & \makecell{Absolute\\Relative Diff.} & \makecell{Squared\\Relative Diff.} & \makecell{Thresh\\($1.25$)} & \makecell{Thresh\\($1.25^2$)} & \makecell{Thresh\\($1.25^3$)}\\
    \hline
    & $\downarrow$ &  $\downarrow$ & $\downarrow$ & $\downarrow$ & $\downarrow$ & $\downarrow$ & $\uparrow$ & $\uparrow$ & $\uparrow$\\
    \Xhline{0.8pt}
    Handcrafted baseline & 0.07045 & 0.7501 & 0.1282 & 0.0886 & 0.1070 & 0.1072 & 0.8907 & \textbf{0.9975} & \textbf{1.0000}\\
    \hline
    \textbf{Ours} & \textbf{0.0189} & \textbf{0.3556} & \textbf{0.0667} & \textbf{0.0443} & \textbf{0.0449} & \textbf{0.0271} & \textbf{0.9890} & 0.9953 & 0.9976\\
    \Xhline{0.8pt}
\end{tabular}
\caption{\hl{Performance of safety envelope estimation in the real-world pedestrian environment under a high latency i.e. slower implementation.}}
\label{table:high-latency}
\end{table*}

\end{document}